\documentclass[lettersize,journal]{IEEEtran}
\usepackage{makecell}
\usepackage{amsmath,amsfonts}
\usepackage{dsfont}
\usepackage{algorithm}
\usepackage{algpseudocode}
\usepackage{hyperref}
\usepackage{cleveref}
\usepackage{array}
\usepackage{courier}
\hypersetup{hidelinks}
\usepackage{threeparttable}
\usepackage{subfigure}
\usepackage{textcomp,soul}
\usepackage{stfloats}
\usepackage{url}
\usepackage{verbatim}
\usepackage{graphicx}
\usepackage{booktabs}
\usepackage{multirow}
\usepackage{eqparbox}
\usepackage{cite}
\usepackage{amsthm}
\usepackage{bm}
\usepackage{amssymb}
\usepackage{xspace}
\usepackage[export]{adjustbox}
\newcolumntype{M}[1]{>{\centering\arraybackslash}p{#1}}

\newcommand{\eg}{\textit{e.g.}\xspace}
\newcommand{\ie}{\textit{i.e.}\xspace}

\newcolumntype{C}[1]{>{\centering\arraybackslash}p{#1}}
\theoremstyle{definition}

\usepackage[table]{xcolor}
\definecolor{lightblue}{rgb}{0.7,0.9,1}
\definecolor{lightgreen}{rgb}{0.7,1,0.7}
\definecolor{newlightgreen}{RGB}{177, 207, 135}
\definecolor{mediumgray}{rgb}{0.55, 0.55, 0.55}
\definecolor{lightgray}{rgb}{0.75, 0.75, 0.75}
\definecolor{lightsalmon}{RGB}{248, 214, 190}
\definecolor{lightpurple}{RGB}{200, 176, 200}

\makeatletter
\pretocmd\@bibitem{\color{black}\csname keycolor#1\endcsname}{}{\fail}
\newcommand\citecolor[1]{\@namedef{keycolor#1}{\color{blue}}}
\makeatother

\newcommand{\inlineicon}[1]{%
    \raisebox{-0.15em}{\includegraphics[height=1em, keepaspectratio=true]{#1}}%
}

\title{
  LUNA-AD: Lightweight Uncertainty-Aware Language Model with Lifelong Learning for Autonomous Driving}
\author{
  Ruoyu Yao, Pei Liu, Ruiguo Zhong, Mingxing Peng, Rui Yang, and Jun Ma, \textit{Senior Member, IEEE}
  \thanks{Ruoyu Yao, Pei Liu, Ruiguo Zhong, Mingxing Peng, Rui Yang, and Jun Ma are with The Hong Kong University of Science and Technology (Guangzhou), China (e-mail: 
  jun.ma@ust.hk).}
  }

\begin{document}
\maketitle

\begin{abstract}
While large language models (LLMs) offer promising reasoning capabilities, their integration into safety-critical driving systems is hindered by limited reasoning diversity, high computational overhead, and static learning paradigms. To address these challenges, we propose LUNA-AD, a lightweight uncertainty-aware language model with lifelong learning for autonomous driving (AD). LUNA-AD features a tri-system architecture that reconciles complex multimodal behavioral reasoning, efficient deployment, and continual refinement. We design a multi-agent analytical system to generate uncertainty-aware decision-making demonstrations through diverse hypothesis exploration. A dual-head lightweight heuristic model is distilled to unify the inference of decision distributions and textual explanations while enabling efficient deployment. Furthermore, a reflection-driven lifelong learning mechanism operates on multimodal decision outputs and preserves strategic diversity, allowing for the refinement of candidate decisions and rationales via closed-loop feedback to enhance driving robustness. Extensive experiments on nuPlan benchmarks demonstrate that LUNA-AD achieves state-of-the-art success rates under both non-reactive and reactive modes, with drastically reduced inference latency compared to existing knowledge-driven AD frameworks.
\end{abstract}


\section{Introduction} \label{sec:intro}
Autonomous driving (AD) is fundamentally a problem of embodied decision-making under uncertainty. Beyond perception and control, the core challenge lies in interpreting complex traffic contexts to generate socially compliant and dynamically feasible behaviors~\cite{schwarting2018planning}. To address this, mainstream architectures typically decompose this task into hierarchical decision-making and planning modules~\cite{hang2020intergfram,jiang2024senna}. While such decomposition offers structural clarity and interpretability, the decision layer is often built upon deterministic rules or learned policies that assume a single optimal behavior for each scene~\cite{peng2025bilevel}. However, real-world traffic is inherently multimodal: at an unsignalized intersection, both yielding and proceeding may be valid depending on subtle social cues and interaction dynamics.

The growing complexity of traffic environments has motivated a shift from rule-based systems toward learning-based cognitive architectures~\cite{chen2024end}. More recently, advances in large language models (LLMs)~\cite{brown2020language,yang2025qwen3} have introduced a new paradigm for reasoning-driven autonomy. Through deliberative reasoning~\cite{wei2022chain} and in-context learning~\cite{brown2020language}, foundation models exhibit interpretable inference traces and knowledge generalization capabilities previously unattainable in existing data-driven frameworks. Their integration into robotics and autonomous driving suggests a promising direction: driving as a knowledge-grounded reasoning problem rather than purely a trajectory optimization task~\cite{Mao2023GPTDriver,shao2024lmdrive}. Specifically, recent efforts leverage these cognitive capabilities to align semantic scene understanding with natural language rationales, significantly enhancing interpretability and contextual awareness~\cite{xu2024drivegpt4,sima2024drivelm}. These developments imply that high-level driving behavior may benefit from foundation-model-based cognition, particularly in long-tail or open-world scenarios where traditional approaches struggle. Despite these advances, a structural inconsistency remains between foundation-model reasoning and embodied AD. Autoregressive language models generate outputs sequentially according to learned probability distributions. In practice, however, standard inference protocols often collapse to a single dominant reasoning trajectory due to deterministic decoding or single-shot sampling~\cite{wangself,yao2023tree}. 
Recent studies reveal that LLMs exhibit path dependence and uncertainty collapse~\cite{bigelowforking}, tending to reinforce early hypotheses even when alternative interpretations are plausible. While this behavior may be acceptable in text generation, it becomes problematic in safety-critical decision-making, where multiple hypotheses must be evaluated simultaneously to account for execution uncertainties~\cite{zhang2021unified,yao2024calmm} or complex interactions~\cite{huang2023gameformer,cheng2024pluto,yao2025hierarchical}.

Parallel to LLM-based reasoning, generative planners based on variational models and diffusion processes~\cite{zheng2024genad,zhengdiffusion} have achieved impressive performance in modeling multimodal trajectory distributions. By sampling diverse motion candidates and selecting via scoring mechanisms, they mitigate trajectory-level mode collapse. Nevertheless, these motion-centric approaches often struggle to align low-level trajectory patterns with high-level traffic semantics, leading to socially inconsistent maneuvers in out-of-distribution scenarios~\cite{jiang2024senna}. To address this, more recent works integrate foundation models with generative planners to incorporate semantic reasoning~\cite{wang2024he,yao2024calmm,fu2025orion,li2025recogdrive,li2025drivevla}. While these hybrid frameworks enhance contextual awareness, they often incur substantial computational overhead due to autoregressive reasoning and large parameter sizes. Scaling laws demonstrate that performance improves with model scale \cite{kaplan2020scaling}, but onboard deployment demands low latency and memory efficiency. Thus, the field faces a deployment paradox: richer reasoning requires larger models, yet real-time driving necessitates lightweight inference.

Knowledge distillation offers a principled mechanism to transfer information from large teacher models to compact student models~\cite{hinton2015distilling}. In the context of general LLMs, recent efforts focus on distilling reasoning chains or instruction-following abilities~\cite{Xu2024Survey}. In autonomous driving, some studies distill multimodal driving strategies to capture uncertainty~\cite{li2024hydra, yu2025distilldrive}, while others compress large reasoning models for reactive decision-making~\cite{mei2024contin,liu2025dsdrive}. Nevertheless, integrating these complementary capabilities remains underexplored. Classical distillation theory emphasizes that soft targets encode “dark knowledge” about inter-class relationships~\cite{hinton2015distilling}. In decision-making for driving, this dark knowledge corresponds to the relative plausibility among multiple behavioral intentions. However, probabilities alone lack the semantic grounding necessary for interpretable decision-making. Bridging large-model cognition and lightweight deployment, therefore, requires a dual-objective transfer mechanism that jointly preserves probabilistic structures and semantic rationales.

Beyond deployment efficiency, another limitation persists in current LLM-based driving systems: their learning paradigm is largely static. While recent works introduce reflection or self-correction mechanisms~\cite{wendilu, mei2024contin, ma2025leapvad}, they typically operate under deterministic decision settings. The deterministic corrections store single outcomes, failing to preserve alternative valid strategies. Thus, lifelong learning must operate on multimodal representations to refine decision patterns and update memory via closed-loop feedback. Together, these observations suggest that scalable AD requires three tightly coupled capabilities: 
\begin{itemize}
    \item Analytical exploration of multimodal decision hypotheses to mitigate reasoning path dependence,
    \item Lightweight deployment jointly preserving probabilistic structures and semantic rationales, and
    \item Lifelong learning mechanisms operating on multimodal decision-making patterns.
\end{itemize}

To address these challenges, we propose \textbf{LUNA-AD}, a \textbf{L}ightweight \textbf{UN}certainty-\textbf{A}ware language model with lifelong learning for \textbf{A}utonomous \textbf{D}riving. 
Compared to our prior work~\cite{yao2026decision} based on static distillation, LUNA-AD introduces a novel tri-system architecture orchestrating deliberative reasoning, efficient deployment, and continual refinement. 
Specifically, an analytical system explores diverse hypotheses to construct uncertainty-aware demonstrations, which are distilled into a lightweight heuristic model for low-latency inference.
Crucially, a progress monitor captures system failures, enabling asynchronous scheduling that decouples high-frequency decision inference from event-triggered textual explanations.
This closed-loop feedback then drives a reflection system to refine candidate decisions and rationales, facilitating continual adaptation.
Finally, this submission provides substantial empirical extensions, including domain transfer studies demonstrating the analytical system's adaptability to multi-view RGB inputs, further corroborated by validation on real-world field-collected data.
Our contributions are summarized as follows:

\begin{itemize}
\item We propose LUNA-AD, a tri-system framework that reconciles complex multimodal reasoning with onboard efficiency. At its core lies a dual-head lightweight language model unifying the inference of decision distributions and textual explanations. Through uncertainty-aware distillation and asynchronous task coordination, the model preserves multimodal semantic rationales while enabling efficient onboard deployment.

\item We introduce a multi-agent analytical reasoning system that generates diverse behavioral hypotheses and aggregates them into uncertainty-aware demonstrations, enabling high-fidelity knowledge distillation and retrieval-augmented learning for the lightweight student model.

\item We propose a reflection-driven lifelong learning mechanism that preserves strategic diversity, utilizing closed-loop feedback to refine decisions and rationales, and continuously update the memory bank for enhanced driving robustness.

\item Extensive experiments demonstrate that LUNA-AD achieves competitive closed-loop driving performance with drastically reduced inference latency against existing knowledge-driven AD frameworks, as evidenced by its state-of-the-art (SOTA) success rates across both non-reactive and reactive modes on the nuPlan~\cite{caesar2021nuplan} Test14-Hard and Test14-Random benchmarks.

\end{itemize}

\section{Related Works} \label{sec:related_works}

\subsection{Decision Making in Autonomous Driving}
Early AD systems predominantly rely on traditional decision-making frameworks, represented by rule-based pipelines~\cite{wei2014behavioral} and game-theoretic models~\cite{hang2020intergfram,huang2024universal}. While offering interpretability and safety guarantees within known scenarios, these methods often struggle with the combinatorial complexity of open-world interactions, rendering them brittle when facing unmodeled traffic patterns~\cite{chen2024end}. To address this, learning-based approaches such as Reinforcement Learning (RL)~\cite{li2021safe,peng2025bilevel}, Inverse RL (IRL)~\cite{huang2023conditional}, and Imitation Learning (IL)~\cite{cheng2024pluto,li2024ego} have gained traction. These approaches optimize policies through environmental interaction or expert mimicry, with recent transformer-based architectures further enhancing scene understanding and multi-agent interaction modeling~\cite{huang2023gameformer,zhengdiffusion}. However, traditional learning-based models often lack explicit reasoning capabilities, thus may fail to generalize to long-tail scenarios~\cite{jiang2024senna}.

\subsection{Large Foundation Models in Autonomous Driving}
The emergence of LLMs has introduced a new paradigm for cognitive autonomy, spurring a line of research that leverage the reasoning and generalization capabilities of foundation models to advance driving decision-making. For instance, GPT-Driver~\cite{Mao2023GPTDriver} explores the use of LLMs for traffic semantic comprehension and action reasoning, while DiLu~\cite{wendilu} and LeapAD~\cite{mei2024contin} further incorporate retrieval augmented generation (RAG) to enable knowledge-driven open-world adaptation. DriveLM stands as a representative work that optimizes the reasoning trajectory by leveraging graph-based visual question answering~\cite{sima2024drivelm}. Recent studies also incorporate foundation models into end-to-end Vision-Language-Action (VLA) frameworks to achieve holistic perception-reasoning integration~\cite{shao2024lmdrive,zhou2025opendrivevla,fu2025orion,li2025recogdrive,li2025drivevla}. Despite these advances, most existing LLM-based driving systems incur substantial memory overhead and suffer from high latency, limiting their deployment on resource-constrained on-board platforms.

\subsection{Efficient Deployment and Lifelong Learning}
To mitigate computational overhead, knowledge distillation transfers capabilities from large teacher models to compact students~\cite{hinton2015distilling}. In autonomous driving, current studies typically bifurcate into uncertainty-aware planning distillation~\cite{li2024hydra,yu2025distilldrive} and reasoning-focused model compression~\cite{mei2024contin,liu2025dsdrive}. This dichotomy hinders the development of lightweight models that simultaneously preserve decision distributions and linguistic explanations. Furthermore, static learning frameworks limit adaptability to environmental distribution shifts~\cite{cheng2024pluto}. While some works introduce reflection mechanisms \cite{wendilu,mei2024contin,ma2025leapvad}, they typically operate under deterministic settings. LUNA-AD advances this direction by combining uncertainty-aware distillation with a reflection-driven lifelong learning mechanism, enabling continuous refinement under closed-loop feedback.

\begin{figure*}
  \centering
    \includegraphics[width=0.9\linewidth]{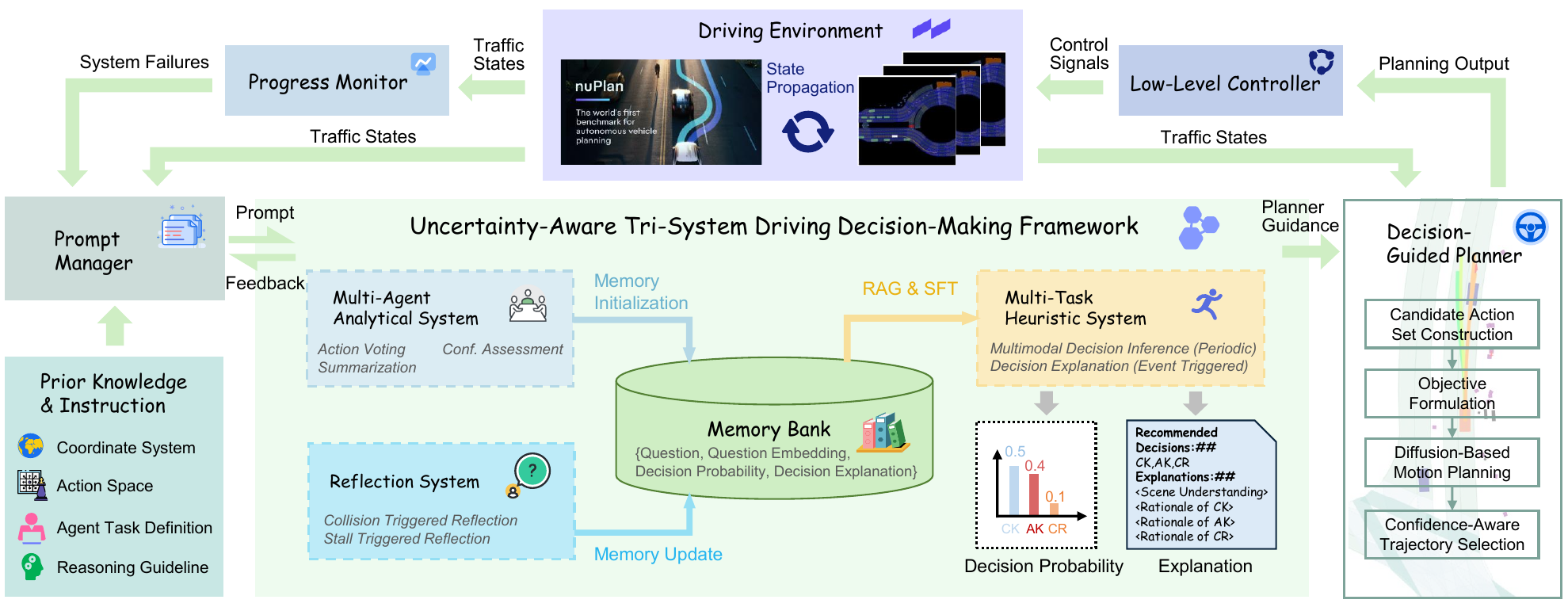}
  \caption{Our methodological architecture: a knowledge-driven tri-system decision-making framework, combined with a prompt manager and a decision-guided planner for closed-loop interaction with the driving environment. The framework includes a \textbf{multi-agent analytical system}, a \textbf{multi-task heuristic system}, and a \textbf{reflection system}, unifying analytical reasoning, fast inference, and lifelong learning. The prompt manager uses predefined logic to generate subsystem messages. The planner produces multimodal trajectories and selects the optimal one for execution, triggering state updates. A progress monitor captures failures (e.g., collisions), enabling the reflection system to refine decisions and update memory.}
  \label{fig:model_framework}
\end{figure*}

\section{Methodology}\label{sec:method}
Our methodological architecture is presented in Fig.~\ref{fig:model_framework}. At its core, we develop a tri-system decision-making framework to reconcile the needs of complex multimodal decision reasoning and inference efficiency, while incorporating lifelong learning capacity. A decision-guided motion planner is utilized to perform multimodal trajectory planning and selection following the decision-making process.

\subsection{Prompt Manager}
The prompt manager dynamically constructs task-specific prompts by selecting and applying an appropriate mapping from a predefined family of prompt-generation functions. This module handles heterogeneous inputs from multiple sources while ensuring each mapping utilizes only relevant information for the current task and subsystem. Below, we formalize its operation.

\textbf{Input Source Spaces.}
The prompt manager receives inputs from five distinct sources:
\begin{itemize}
    \item \(\mathcal{K}\): prior knowledge and instruction,
    \item \(\mathcal{S}_{\text{traf}}\): traffic states in the driving environment,
    \item \(\mathcal{S}_{\text{fail}}\): system failures captured by the progress monitor,
    \item \(\mathcal{F}\): linguistic feedback from subsystems, and
    \item \(\mathcal{\tilde{Q}}\): queries generated by the prompt manager at previous mappings.
\end{itemize}
We specify these concepts in \Cref{sec:sys analytical,sec:sys heuristic,sec:sys reflect}.

\textbf{Prompt-Generation Mappings.}
Let $\{\mathcal{P}_{j}\}_{j \in J}$ be the family of prompt-generation mappings. Each mapping $\mathcal{P}_j$ is defined over a subset of input sources:
\begin{align}
  &\text{Dependency set:} & D_j &\subseteq \{\mathcal{K},\mathcal{S}_{\text{traf}},\mathcal{S}_{\text{fail}}, \mathcal{F}, \mathcal{\tilde{Q}}\}, \label{eq:depset} \\
  &\text{Input domain:} & \mathcal{D}_j &= \prod_{\mathcal{X} \in D_j} \mathcal{X}, \label{eq:domain} \\
  &\text{Mapping function:} & \mathcal{P}_j &: \mathcal{D}_j \to \mathcal{Q}. \label{eq:mapping}
\end{align}
Here \(\mathcal{Q}\) denotes the space of queries generated at the current mapping, a subspace of natural language space: \(\mathcal{Q} \subseteq \mathcal{L}_{\text{nat}}\).

\textbf{Task-Specific Mapping Selection.}
The prompt manager selects prompt-generation mappings based on task type. Specifically, five types of tasks with mappings are incorporated into our framework:
\begin{itemize}
    \item \(\mathcal{P}_{\text{av}}\): action voting,
    \item \(\mathcal{P}_{\text{ca}}\): confidence assessment,
    \item \(\mathcal{P}_{\text{sum}}\): summarization,
    \item \(\mathcal{P}_{\text{md}}\): multimodal decision-making, and
    \item \(\mathcal{P}_{\text{refl}}\): reflection.
\end{itemize}
During initial memory collection, the multi-agent analytical system conducts action voting, confidence assessment, and summarization utilizing mappings \(\{\mathcal{P}_{\text{av}}, \mathcal{P}_{\text{ca}}, \mathcal{P}_{\text{sum}}\}\). Following that, the operation \(\mathcal{P}_{\text{md}}\) is leveraged to query the multi-task heuristic system for multimodal decision-making in both training and inference. As system failures are captured in closed-loop driving, a reasoning task is assigned to the reflection system with the invocation of \(\mathcal{P}_{\text{refl}}\). We explain details regarding the prompt-generation mappings and subsystem operations in \Cref{sec:sys analytical,sec:sys heuristic,sec:sys reflect}.

\subsection{Multi-Agent Analytical System for Uncertainty-Aware Decision-Making Demonstrations}\label{sec:sys analytical}

\begin{table}[t]
    \centering
    \caption{Elements included in traffic state representation.}
    \begin{tabular}{l|l}
        \toprule
        \textbf{Element} & \textbf{Content} \\ 
        \midrule
         \multirow{14}{*}{\(\mathcal{R}_t\)} & Based on the category of road section:\\ 
         & \(\bullet\) Normal Multi-Lane: \\
         & \(\quad \quad \mathcal{R}_t=\langle N_{\text{c}}, j_{\text{c}} \rangle\),\\
         & \(\bullet\) Multi-Lane Near Junction: \\
         & \(\quad \quad \mathcal{R}_t=\langle N_{\text{c}}, j_{\text{c}}, \{I_{j_{\text{c}}}\}^{N_{\text{c}}}_{j_{\text{c}}}, \{\mathbf{p}^{\text{exit}}_{j_{\text{g}}}\}^{N_{\text{g}}}_{j_{\text{g}}=1}\rangle\),\\
         & \(\bullet\) Junction: \\
         & \(\quad \quad \mathcal{R}_t=\langle \{\mathbf{p}^{\text{exit}}_{j_{\text{g}}}\}^{N_{\text{g}}}_{j_{\text{g}}=1} \rangle\),\\
         & where \\
         & \(N_{\text{c}}\): number of lanes in the current road,\\
         & \(N_{\text{g}}\): number of lanes in the target road,\\
         & \(j_{\text{c}}\): index of the current lane (left to right), \\
         & \(j_{\text{g}}\): index of the target lane (left to right), \\
         & \(I_{j_{\text{c}}}\): boolean indicating if lane \(j_{\text{c}}\) connects to the target, \\
         & \(\mathbf{p}^{\text{exit}}_{j_{\text{g}}}\): position of junction exit point on the target road. \\
         \midrule
         \multirow{13}{*}{\(\mathbf{s}^{\text{veh}}_{i_\text{v},t}, \mathbf{s}^{\text{vru}}_{i_\text{r},t}, \mathbf{s}^{\text{stat}}_{i_\text{s},t}\)} & Based on the type of the road object:\\
         & \(\bullet\) Vehicle: \\
         & \(\quad \quad \mathbf{s}^{\text{veh}}_{i_\text{v},t}=\langle \mathbf{p}^{\text{veh}}_{i_{\text{v}},t}, \psi^{\text{veh}}_{i_{\text{v}}, t}, v^{\text{veh}}_{i_{\text{v}}, t}, l^{\text{veh}}_{i_{\text{v}}}, w^{\text{veh}}_{i_{\text{v}}}\rangle\), \\
         & \(\bullet\) VRU: \\
         & \(\quad \quad \mathbf{s}^{\text{vru}}_{i_\text{r},t}=\langle \mathbf{p}^{\text{vru}}_{i_{\text{r}},t}, \psi^{\text{vru}}_{i_{\text{r}}, t}, v^{\text{vru}}_{i_{\text{r}}, t}, l^{\text{vru}}_{i_{\text{r}}}, w^{\text{vru}}_{i_{\text{r}}}\rangle\), \\
         & \(\bullet\) Static Object: \\
         & \(\quad \quad \mathbf{s}^{\text{stat}}_{i_\text{s},t}=\langle \mathbf{p}^{\text{stat}}_{i_{\text{s}},t}, \psi^{\text{stat}}_{i_{\text{s}}, t}, l^{\text{stat}}_{i_{\text{s}}}, w^{\text{stat}}_{i_{\text{s}}}\rangle\),\\
         & where \\
         & \(\{\mathbf{p}^{\text{veh}}_{i_{\text{v}},t}, \mathbf{p}^{\text{vru}}_{i_{\text{r}},t}, \mathbf{p}^{\text{stat}}_{i_{\text{s}},t}\}\): position at time-step \(t\),\\
         & \(\{\psi^{\text{veh}}_{i_{\text{v}},t}, \psi^{\text{vru}}_{i_{\text{r}},t}, \psi^{\text{stat}}_{i_{\text{s}},t}\}\): heading angle at time-step \(t\),\\
         & \(\{v^{\text{veh}}_{i_{\text{v}},t}, v^{\text{vru}}_{i_{\text{r}},t}\}\): speed at time-step \(t\),\\
         & \(\{l^{\text{veh}}_{i_\text{v}}, l^{\text{vru}}_{i_\text{r}}, l^{\text{stat}}_{i_\text{s}}\}\): length of bounding box, \\
         & \(\{w^{\text{veh}}_{i_\text{v}}, w^{\text{vru}}_{i_\text{r}}, w^{\text{stat}}_{i_\text{s}}\}\): width of bounding box. \\
         \midrule
         \(\lambda_t\) & \(\bullet\) \(\quad \lambda_t \in \{\text{Yellow},\text{Red},\text{Green},\text{None}\}\) \\
         \midrule
         \multirow{5}{*}{\(\kappa_t\)} & Based on the category of road section:\\
         & \(\bullet\) Normal Multi-Lane:\\
         & \(\quad \quad \kappa_t \in \{\text{None}\}\),\\
         & \(\bullet\) Multi-Lane Near Junction or Junction:\\
         & \(\quad \quad \kappa_t \in \{\text{Left},\text{Right},\text{Straight}\}\). \\
        \bottomrule
    \end{tabular}
    \label{tab:traffic states}
\end{table}

\begin{figure}[t]
  \centering
   \includegraphics[width=1\linewidth]{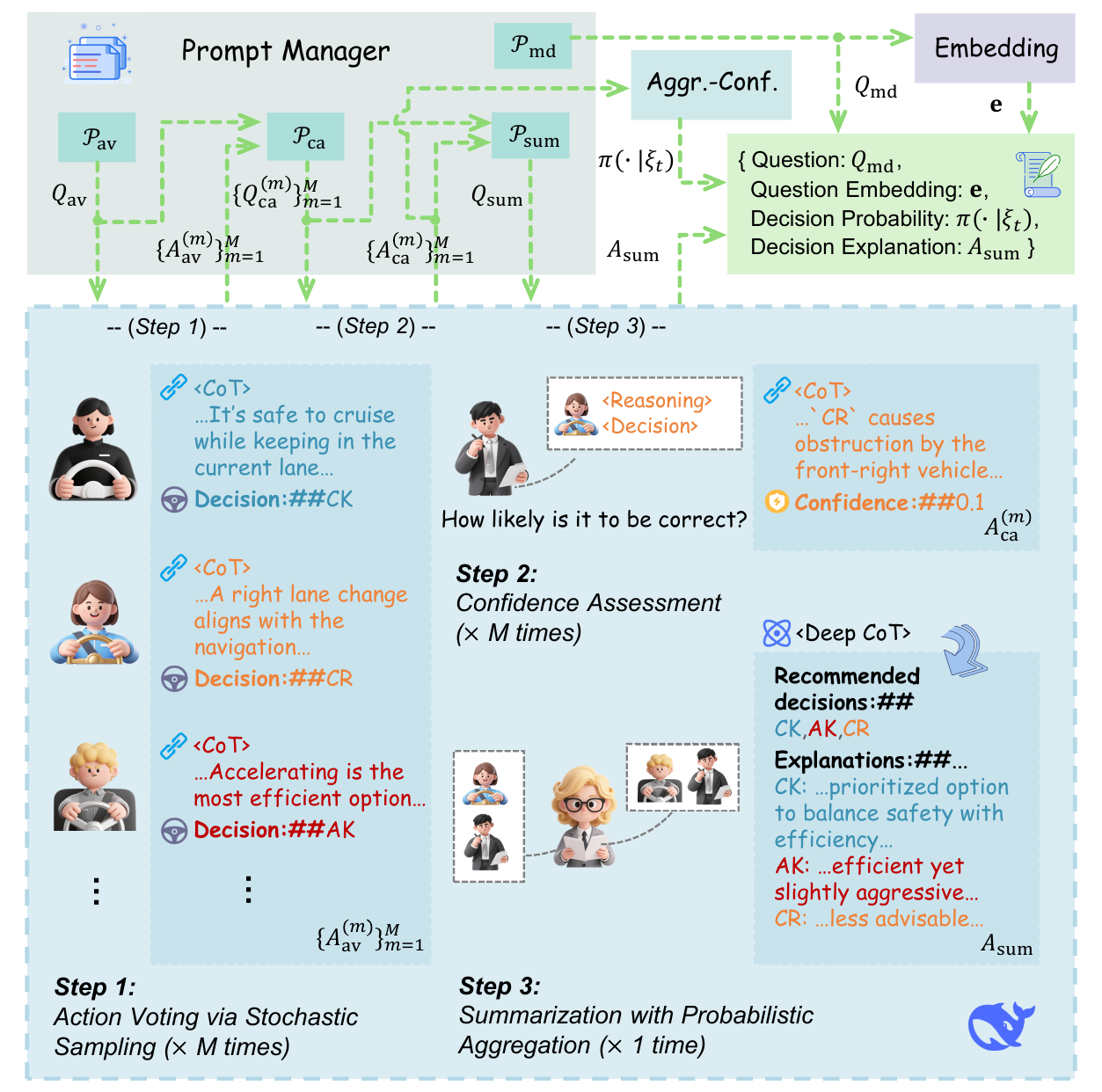}
   \caption{Generation of uncertainty-aware decision-making demonstrations by the multi-agent analytical system. The analytical system interacts with the prompt manager to complete a three-stage collaborative reasoning: \textbf{action voting}, \textbf{confidence assessment}, and \textbf{summarization}. This produces multimodal decision probabilities with explanation, recorded as a memory item alongside the multimodal decision query and embedding.}
   \label{fig:initial memory collection}
\end{figure}

Conventional knowledge distillation approaches for AD typically rely on single-pass inference from foundation models, which suffer from limited reasoning diversity~\cite{wangself}, the tendency to commit prematurely to a single hypothesis without adequate exploration of the multimodal intention space~\cite{yao2024calmm}. To overcome this limitation, we formalize decision-making as a stochastic exploration process over the action manifold, where diverse hypotheses are generated, critically evaluated, and probabilistically aggregated.

\subsubsection{Scene Representation and Reasoning Initialization} 
At time-step $t$, the traffic states are encoded as a structured scene descriptor $\xi_t \in \mathcal{S}_{\text{traf}}$, comprising:
\begin{equation}
    \xi_t = \big\langle \mathcal{R}_t, \{ \mathbf{s}^{\text{veh}}_{i_\text{v},t} \}_{i_\text{v}=1}^{N_{\text{v}}}, \{ \mathbf{s}^{\text{vru}}_{i_\text{r},t} \}_{i_{\text{r}}=1}^{N_{\text{r}}}, \{ \mathbf{s}^{\text{stat}}_{i_{\text{s}},t} \}_{i_{\text{s}}=1}^{N_\text{s}}, \lambda_t, \kappa_t \big\rangle,
    \label{eq:scene_descriptor}
\end{equation}
where $\mathcal{R}_t$ denotes road topology. $\mathbf{s}^{\text{veh}}_{i_\text{v},t}$, $\mathbf{s}^{\text{vru}}_{i_\text{r},t}$, and $\mathbf{s}^{\text{stat}}_{i_{\text{s}},t}$  represent states of surrounding vehicles, vulnerable road users (VRUs) and static objects. Specifically, we use \(i_v=0\) to index the state of the ego vehicle. $\lambda_t$ encodes traffic light phases, and $\kappa_t$ specifies navigation information. We detail the content of each element of the descriptor in \Cref{tab:traffic states}.

Meanwhile, we inject prior knowledge and instruction via natural language description to initialize the reasoning process of each agent. This includes the explanation of \textbf{coordinate system}, \textbf{action space}, \textbf{agent task}, and \textbf{reasoning guideline}. Following \cite{yao2024calmm}, we define a polar coordinate system \(\mathcal{B}\) centered at the ego-vehicle’s geometric center, with zero azimuth aligned with the vehicle’s heading direction, to describe the distance and azimuth of observed objects. The action space \(\mathcal{A}\) is a Cartesian product of longitudinal and lateral available actions, with \(\mathcal{A}_{\text{lon}} = \{\text{Acceleration}, \text{Deceleration}, \text{Cruise}, \text{Stop}\}\) and \(\mathcal{A}_{\text{lat}} = \{\text{Left Lane-Change}, \text{Right Lane-Change}, \text{Lane Keep}\}\). While the definitions of these basic concepts remain consistent across all prompts, the descriptions of agent-specific tasks and reasoning guidelines are tailored to the distinct roles. We denote the basic concepts augmented by the instructions specialized for agent/task \(j\) as \(\mathbf{K}_{j}(\mathcal{B},\mathcal{A}) \in \mathcal{K}\).

\begin{table*}[htbp]
\centering
\resizebox{\textwidth}{!}{%
\begin{minipage}{\textwidth}
\caption{Key portions of the agent-specific instructions for action voting, confidence assessment, summarization, and reflection agents.}
\label{tab:agent-specific instruction}
\renewcommand{\arraystretch}{1.0} 
\begin{tabular}{>{}p{3.0cm}p{\dimexpr\textwidth-3.0cm-4\tabcolsep}}
\toprule
\multicolumn{1}{l}{\textbf{Agent Task}} & \multicolumn{1}{l}{\textbf{Instruction Message}} \\
\midrule
\includegraphics[height=1.8em, valign=c]{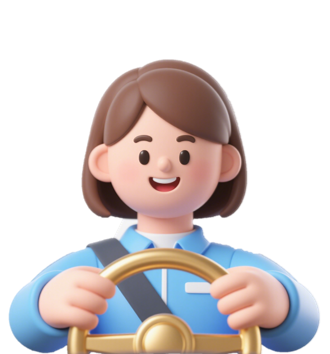}%
\hspace{0.3em} Action Voting & 
\textbf{Step 1: Holistic Scene Understanding}. Analyze the driving context, including the road structure, traffic light, navigation for the ego vehicle (if applicable), and the movement of surrounding vehicles, and describe their implications on the ego vehicle's behavior. \newline 
\textbf{Step 2: Key Object Identification}. Identify the most important objects affecting the ego vehicle's behavior (vehicles, VRUs, or static objects). Focus only on critical ones, such as: vehicles with lane conflicts, \eg, vehicles in the same lane, vehicles intending to switch into the ego lane, vehicles in the lane the ego vehicle will switch to; vehicles at junctions and close ahead of the ego vehicle; VRUs within 15 m, with a speed and heading angle suggesting potential collision; Static objects directly obstructing the ego vehicle's path. \newline 
\textbf{Step 3: Action Selection}: Based on the step 1 \& 2 analysis and common-sense knowledge, predict the best decision from the \{action\_set\}.\\
\midrule
\includegraphics[height=1.8em, valign=c]{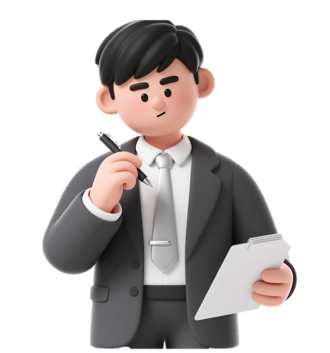}%
\hspace{0.3em}
Conf. Assessment & 
How likely is the above decision-making process to be correct? Analyze the response, provide your reasoning concisely, and give your confidence (between 0.0 and 1.0) in this response.\\
\midrule
\includegraphics[height=1.8em, valign=c]{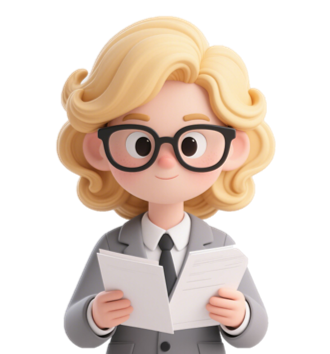}%
\hspace{0.3em}
Summarization & 
You should take into account the responses of both the action voting agents and the confidence assessment agent, then conclude all types of observed decisions and their rationales. \newline
- The overall probability of a decision \(D\) is calculated by: \(\mathbf{Pr}(D) = \sum^M_{m=1} \mathds{1}{(d_m=D)} \cdot c_m / \sum^M_{{m=1}} c_m\), where \(d_m\) denotes the decision proposed by the \(m\)-th action voting agent and \(c_m\) denotes the confidence estimate.
\newline
- In your response content, make sure to sort the observed decisions in descending order based on their probabilities.
\newline
- The rationale of a decision should include the associated analysis of key objects in the scene, and the reason why the driving strategy is advisable.
\newline
- You may incorporate confidence-related analysis internally during reasoning, but must not leak the specific confidence scores in your response.
\newline
- You should not leak any information about other agents or the workflow of this task. Frame your driving recommendations as if they result entirely from your own reasoning.\\
\midrule
\includegraphics[height=1.8em, valign=c]{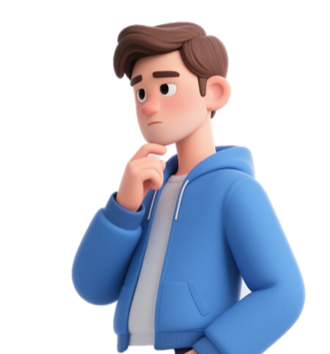}%
\hspace{0.3em}
Reflection & 
Based on the question sent to the decision-maker, the decision-maker's response, and the outcome caused by the executed decision, analyze the driving situation and revise the decision-making process if necessary.\newline
- Try to identify unreasonable aspects in the decision-making process (\eg, flawed priorities, invalid rationales), and provide a corrected response. If there is no collision and the decision leading to conservative driving is justified, output exactly: `normal' — nothing else.
\newline
- Your reflection should be conducted internally as part of your reasoning, and your final response must follow exactly the same format and structure as the original answer.
\newline
- You must not leak any information about the decision-maker or reveal the fact that this is a reflection task.
\newline
- The proposed decisions must be selected exclusively from the action set specified in the original question.\\
\bottomrule
\end{tabular}
\end{minipage}%
}
\end{table*}

\subsubsection{Three-Stage Collaborative Reasoning} 
As shown in Fig.~\ref{fig:initial memory collection}, the analytical process unfolds through three functionally specialized stages that collectively implement \textbf{action voting} $\rightarrow$ \textbf{confidence assessment} $\rightarrow$ \textbf{summarization}. This decomposition reframes the complex multimodal driving decision-making problem as a set of tractable, well-defined subtasks, each delegated to a specialized agent for effective resolution.

\textbf{Stage 1: Action Voting via Stochastic Sampling.} We instantiate an ensemble of $M$ independent reasoning agents by prompting a foundation model with identical queries under non-deterministic decoding, using Top-K sampling with a temperature parameter \(\widetilde{\tau}>0\). The query generated by the prompt manager is expressed by:
\begin{equation}
    Q_{\text{av}} = \mathcal{P}_{\text{av}}\left(\xi_t, \mathbf{K}_{\text{av}}\left(\mathcal{B},\mathcal{A}\right)\right).
\end{equation}
Here \(\mathbf{K}_{\text{av}}\left(\mathcal{B},\mathcal{A}\right)\) instructs action voting agents to follow a reasoning guideline of three explicit steps to derive the action, encompassing holistic scene understanding, key object identification, and action selection.
We exhibit details regarding this guideline in the first row of \Cref{tab:agent-specific instruction}.
Then, each agent produces a complete reasoning trajectory:
\begin{subequations}
    \begin{align}
    \phi^{(m)}_1 &= \mathcal{G}_1 \big( Q_{\text{av}}; \widetilde{\tau} \big), \label{eq:stage1_1} \\
    \phi^{(m)}_2 &= \mathcal{G}_2 \big( Q_{\text{av}}, \phi^{(m)}_1; \widetilde{\tau} \big), \label{eq:stage1_2} \\
    \phi^{(m)}_3 &= \mathcal{G}_3 \big( Q_{\text{av}}, \phi^{(m)}_1, \phi^{(m)}_2; \widetilde{\tau} \big), \label{eq:stage1_3}\\
    A^{(m)}_{\text{av}} &= \phi^{(m)}_1 \Vert \phi^{(m)}_2 \Vert \phi^{(m)}_3,
\end{align}
\end{subequations}
where \(\Vert\) denotes the concatenation operator. \(m \in \{1,...,M\}\) indexes the action voting agents. \(\mathcal{G}_1\),  \(\mathcal{G}_2\), and \(\mathcal{G}_3\) represent the autoregressive text generation in structured Chain-of-Thought (CoT) reasoning steps, with the generated contents \(\mathcal{\phi}^{(m)}_1\), \(\mathcal{\phi}^{(m)}_2\), and \(\mathcal{\phi}^{(m)}_3\) constituting the complete response \(A^{(m)}_{\text{av}}\). The final action is extracted via deterministic parsing $\mathcal{M}_{\text{act}}$: $\mathbf{a}_t^{(m)} = \mathcal{M}_{\text{act}}\big(\phi^{(m)}_3\big) \in \mathcal{A}$. This stage essentially implements \(M\) independent unimodal decision-making processes, obtaining diverse hypotheses over valid driving strategies.

\textbf{Stage 2: Confidence Assessment.} For each hypothesis proposed by action voting agents, we instantiate a confidence assessment agent to analyze the response and estimate its correctness. This derives confidence scores for the response of each voting agent, effectively suppressing low-quality or hallucinated outputs while avoiding entrenchment in a single reasoning trajectory~\cite{xiongcan}. 

The prompt manager synthesizes the query \(Q_{\text{av}}\) generated in action voting with the feedback \(A^{(m)}_{\text{av}}\) to construct a query for the confidence assessment agent:
\begin{equation}
     Q^{(m)}_{\text{ca}} = \mathcal{P}_{\text{ca}}\left(Q_{\text{av}}, A^{(m)}_{\text{av}}, \mathbf{K}_{\text{ca}}\left(\mathcal{B},\mathcal{A}\right)\right),
\end{equation}
where \(\mathbf{K}_{\text{ca}}\left(\mathcal{B},\mathcal{A}\right)\) incorporates the instruction shown in the second row of \Cref{tab:agent-specific instruction} to elicit verbalized confidence.
We formalize the confidence assessment process for the \(m\)-th voting agent as:
\begin{subequations}
    \begin{align}
        A^{(m)}_{\text{ca}} &= \mathcal{G}_4( Q^{(m)}_{\text{ca}};\underline{\tau} ),
    \label{eq:confidence}\\
    c^{(m)} &= \mathcal{M}_{\text{conf}}(A^{(m)}_{\text{ca}})\in [0,1],
    \end{align}
\end{subequations}
where \(\mathcal{G}_4\) represents the text generation in confidence assessment. We use a temperature parameter \(\underline{\tau}=0\) to ensure deterministic decoding in the evaluation task. \(\mathcal{M}_{\text{conf}}\) denotes a deterministic parsing to extract confidence value \(c^{(m)}\) from the response \(A^{(m)}_{\text{ca}}\).
Critically, evaluation occurs in isolation for each hypothesis without exposure to other agents' outputs, which prevents anchoring bias induced by majority opinions~\cite{zhu2025conformity}.

\textbf{Stage 3: Summarization with Probabilistic Aggregation.} Recall that the objective of the analytical system is to generate multimodal decision-making demonstrations, which incorporate outputs regarding both textual explanations and probabilistic distributions. We utilize a summarization agent to comprehend the complete question-answering records in the above stages, aggregating diverse reasoning trajectories into a compact description of recommended decisions and rationales:
\begin{subequations}
    \begin{align}
        Q_{\text{sum}} &= \mathcal{P}_{\text{sum}}\left(Q^{(1:M)}_{\text{ca}}, A^{(1:M)}_{\text{ca}}, \mathbf{K}_{\text{sum}}(\mathcal{B},\mathcal{A})\right),\\
        A_{\text{sum}} &= \mathcal{G}_5(Q_{\text{sum}};\underline{\tau}).
    \end{align}
\end{subequations}
The key portion of the instruction message for the summarization agent is exhibited in the third row of \Cref{tab:agent-specific instruction}. Notice that the summarization agent is designed to conceal the presence of other agents and internal workflow details, thereby compelling the student model to capture multimodal decision-making patterns through individualized reasoning.

Concurrently, we aggregate the confidence scores obtained via the above stages to estimate the decision distribution conditioned on traffic states:
\begin{equation}
    \pi(\mathbf{a} \mid \xi_t) = \frac{\sum_{m=1}^M c^{(m)} \cdot \mathds{1}\big( \mathbf{a}^{(m)}_t = \mathbf{a} \big)}{\sum^M_{m=1} c^{(m)}}, \ \forall \mathbf{a} \in \mathcal{A},
    \label{eq:probabilistic_synthesis}
\end{equation}
where \(\mathds{1}(\cdot)\) represents the indicator function.

\subsubsection{Memory Accumulation}
Finally, we store the decision-making demonstrations generated by the analytical system as memory items used by the student model. This requires the construction of  a multimodal decision-making query:
\begin{equation}
    Q_{\text{md}} = \mathcal{P}_{\text{md}}\left(\xi_t, \mathbf{K}_{\text{md}}(\mathcal{B},\mathcal{A})\right).
\end{equation}
Here \(\mathbf{K}_{\text{md}}(\mathcal{B},\mathcal{A})\) does not incorporate an elaborate reasoning guideline like that of \(\mathbf{K}_{\text{av}}(\mathcal{B},\mathcal{A})\), since the student model is developed to automatically capture multimodal decision-making patterns via supervised fine-tuning (SFT) and RAG. Then, an embedding operation maps the portion of the traffic state description in \(Q_{\text{md}}\) to a high-dimensional vector \(\mathbf{e}\), enabling the retrieval of similar driving scenes in RAG. Incorporating the textual explanation and decision distribution, the complete memory item is represented by:
\begin{equation}
    \upsilon(\xi_t) = \big\langle Q_{\text{md}}, \; \mathbf{e}, \; \pi(\cdot \mid \xi_t), \; A_{\text{sum}} \big\rangle.
    \label{eq:demonstration_tuple}
\end{equation}
We then index each item using \(n \in \{1,...,N\}\), building the memory bank \(\Upsilon=\{ \upsilon^{(n)} \}^N_{n=1}\) for retrieval purposes.

\subsection{Multi-Task Heuristic System for Lightweight Deployment}\label{sec:sys heuristic}

\begin{figure}[t]
  \centering
   \includegraphics[width=0.97\linewidth]{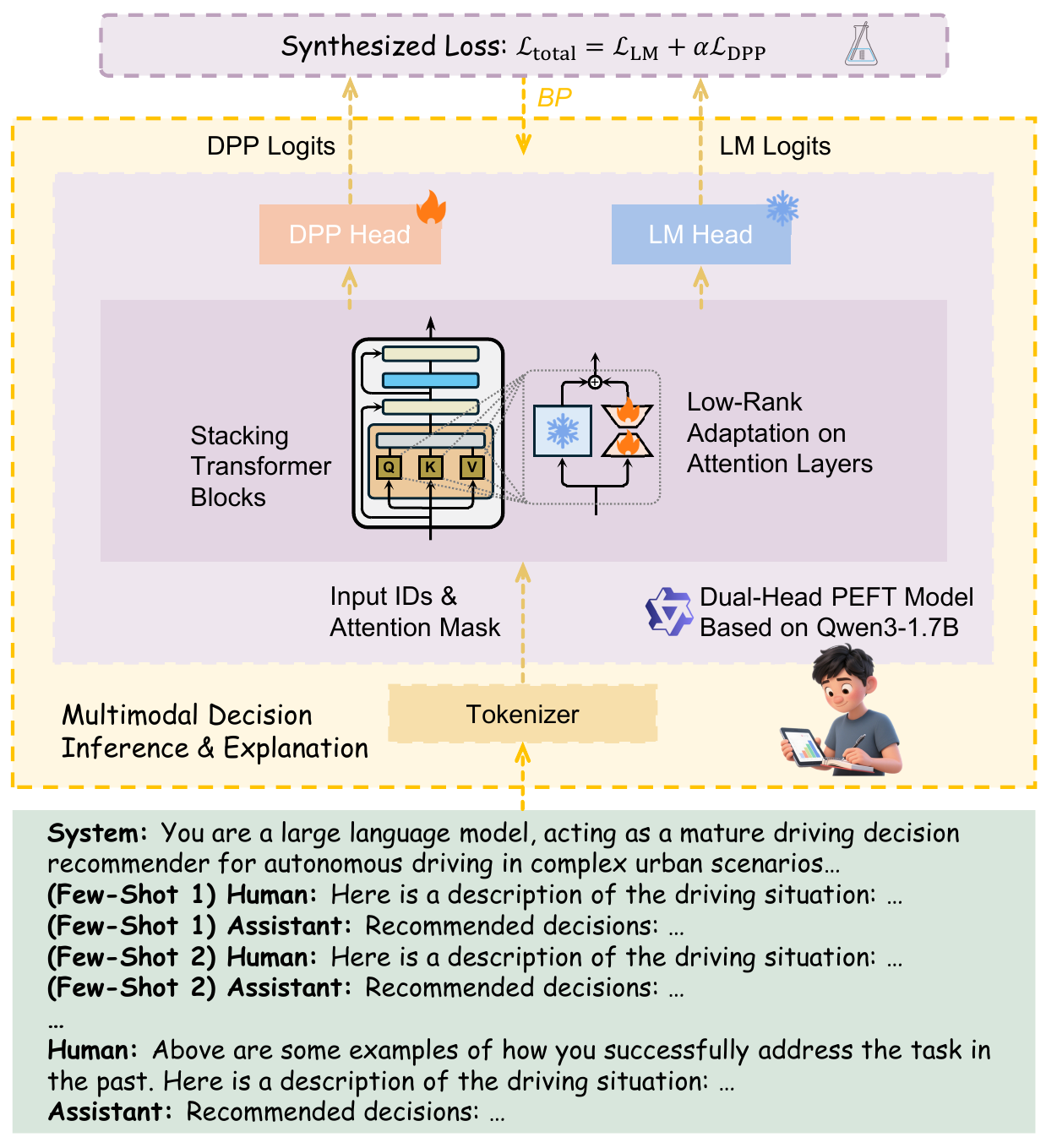}
   \caption{An illustration of the dual-head lightweight architecture for decision probability prediction (DPP) and language modeling (LM). The complementary objectives are synthesized in parameter-efficient fine-tuning (PEFT) to facilitate the understanding of multimodal decision-making patterns.}
   \label{fig:sft}
\end{figure}

The heuristic system bridges the chasm between deliberative reasoning quality and on-board deployment constraints through a dual-objective knowledge transfer mechanism that preserves both the probabilistic structure of multimodal intentions and their semantic rationales, following the slow-fast cognitive paradigm~\cite{mei2024contin}.

\subsubsection{Retrieval-Augmented Context Construction}
Given a query scene with embedding \(\mathbf{e}\), we retrieve $K$ contextually relevant demonstrations $\{ \upsilon^{(n_{k})} \}$ from the memory bank \(\Upsilon\) via cosine embedding similarity.
During training, we sample $K$ stochastically from a discrete uniform distribution $K \sim \mathcal{U}( \{K_{\min}, \dots, K_{\max}\})$, which exposes the model to variable context lengths to enhance robustness to retrieval variability during deployment. The augmented query sequence becomes:
\begin{equation}
    \bar{Q}_{\text{md}} = \Big(\big\Vert_{k=1}^K \Big[ Q^{(n_k)}_{\text{md}} \Vert A^{(n_k)}_{\text{sum}} \Big] \Big) \;\Vert\; Q_{\text{md}},
    \label{eq:augmented_query}
\end{equation}
enabling the lightweight model to perform in-context learning typically reserved for large foundation models~\cite{wendilu}.

\subsubsection{Dual-Head Architecture with Uncertainty-Aware Distillation}
As shown in Fig.~\ref{fig:sft}, we present a dual-head architecture based on a lightweight language model to simultaneously distill the capabilities of decision probability prediction and textual rationale generation. Taking the complete dialog \(\bar{Q}_{\text{md}} \Vert A_{\text{sum}}\) as input, the model learns two complementary objectives through parameter-efficient fine-tuning.

We denote the tokenized input sequence by:
\begin{equation}
    \mathbf{x} = \text{Tokenize}(\bar{Q}_{\text{md}} \Vert A_{\text{sum}}) = [x_1, x_2, \dots, x_I],
\end{equation} 
where \(I\) is the length of the sequence. The sequence is processed by the transformer blocks of the language model \(f_{\theta}\) to generate hidden states \(\mathbf{H}=[\mathbf{h}_1, \mathbf{h}_2, \dots, \mathbf{h}_I]\). We identify the \textit{decision commitment position} $i^*$, the token immediately preceding the assistant's response onset, where high-level decision probabilities should be emitted.

The model learns two complementary objectives through a synthesized loss:

\textbf{Language Modeling for Decision Explanation.} To facilitate comprehension of the entire dialog context, we have the model perform next-token prediction over the given sequence, then compute the loss across all tokens in the sequence~\cite{wang2022language}:
\begin{equation}
        \mathcal{L}_{\text{LM}} = -\frac{1}{I} \sum^{I}_{i=1} \log p\big(x_{i+1}|\mathbf{h}_{1:i};\theta_{\text{LM}}\big),
    \label{eq:lm_loss}
\end{equation}
where \(\theta_{\text{LM}}\) denotes the parameters of the LM head.

\textbf{Decision Probability Prediction.} We employ a lightweight classifier head $f_{\theta_{\text{DPP}}}$ with softmax function to map the commitment state $\mathbf{h}_{i^*}$ to a distribution $\hat{\pi}(\cdot \mid \xi_t)$ over $\mathcal{A}$:
\begin{equation}
    \hat{\pi}(\mathbf{a} \mid \xi_t) 
    = \frac{
        \exp\big( [f_{\theta_{\text{DPP}}}(\mathbf{h}_{i^*})]_{\mathbf{a}} \big)
    }{
        \sum_{\mathbf{a}' \in \mathcal{A}} 
        \exp\big( [f_{\theta_{\text{DPP}}}(\mathbf{h}_{i^*})]_{\mathbf{a}'} \big)
    },
    \quad \forall \mathbf{a} \in \mathcal{A}.
    \label{eq:decision_prediction}
\end{equation}
The model is trained via KL divergence to preserve the uncertainty structure from analytical demonstrations in Eq.~\eqref{eq:probabilistic_synthesis}:
\begin{equation}
    \mathcal{L}_{\text{DPP}} 
    = \sum_{\mathbf{a} \in \mathcal{A}} 
      \pi(\mathbf{a} \mid \xi_t) 
      \log \frac{\pi(\mathbf{a} \mid \xi_t)}{\hat{\pi}(\mathbf{a} \mid \xi_t)}.
    \label{eq:dpp_loss}
\end{equation}
 Incorporating a balancing factor \(\alpha\), the complete training objective is constructed to enable the joint optimization of the dual tasks:
 \begin{equation}
     \mathcal{L}_{\text{total}} = \mathcal{L}_{\text{LM}} + \alpha \mathcal{L}_{\text{DPP}}.
 \end{equation}

\subsubsection{Dual-Task Coordination in Inference Time}
To coordinate the dual tasks during inference, we have the heuristic system \textit{periodically} perform decision probability prediction while generating decision explanations in an \textit{event-triggered} fashion. This allows the decision-making framework to send probabilistic guidance signals to the downstream planner in regular intervals, avoiding time-consuming autoregressive text generation when it is not required. We explain the decision-guided motion planning process in~\Cref{sec:motion planner}, and we detail the event-triggered decision explanation that closely connects to the reflection system in~\Cref{sec:sys reflect}.

\subsection{Decision-Guided Motion Planning}\label{sec:motion planner}
The motion planner translates high-level probabilistic decisions into executable trajectories while respecting the uncertainty quantified by the heuristic system. Rather than treating decisions as deterministic commands, we formalize planning as an uncertainty-aware optimization problem where decision probability directly modulates trajectory generation complexity, following the framework established in~\cite{yao2024calmm}.

\textbf{Adaptive Candidate Set Construction.} Given predicted distribution $\hat{\pi}(\cdot \mid \xi_t)$, we construct a probability-thresholded candidate action set:
\begin{equation}
    \mathcal{C}_t = \big\{ \mathbf{a} \in \mathcal{A} \mid \hat{\pi}(\mathbf{a} \mid \xi_t) \geq \gamma_p \big\},
    \label{eq:candidate_set}
\end{equation}
where $\gamma_p$ serves as a tunable threshold. In contrast to fixed-candidate approaches~\cite{yao2024calmm}, this mechanism yields adaptive planning complexity, generating compact candidate sets in deterministic scenarios to reduce computational load while expanding coverage in ambiguous situations to preserve multimodality.

\textbf{Trajectory Generation and Refinement.} For each $\mathbf{a} \in \mathcal{C}_t$, we define a trajectory optimization objective balancing decision adherence and general quality:
\begin{equation}
    \mathcal{J}_{\mathbf{a}} = \big( \mathcal{J}_{\mathbf{a}}^{\text{dec}} \big)^{\omega_{\text{dec}}} \cdot \big( \mathcal{J}^{\text{gen}} \big)^{\omega_{\text{gen}}},
    \label{eq:objective}
\end{equation}
where $\mathcal{J}_{\mathbf{a}}^{\text{dec}}$ penalizes deviations from action-specific lane/speed targets, identical to the definition in~\cite{yao2024calmm}. $\mathcal{J}^{\text{gen}}$ instantiates the Predictive Driver Model (PDM) scorer~\cite{dauner2023parting} evaluating safety, comfort, and traffic rule compliance. \(\omega_{\text{dec}}, \omega_{\text{gen}} \, \in \mathbb{R}_{+}\) serve as balancing parameters.

Under each constructed objective \(\mathcal{J}_{\mathbf{a}}, \, \forall \mathbf{a} \in \mathcal{C}_t\), trajectory proposals are generated and iteratively refined via gradient-free diffusion optimization proposed by \textit{Diffusion-ES}~\cite{yang2024diffusion}:
\begin{subequations}
    \begin{align}
    \mathcal{X}_{\mathbf{a}} &= \text{Diffusion-ES}(\mathcal{J}_\mathbf{a}), \label{eq:diffusion_gen} \\
    \mathbf{X}^*_\mathbf{a} &= \arg\max_{\mathbf{X}_{\mathbf{a}} \in \mathcal{X}_a} \mathcal{J}_{\mathbf{a}}(\mathbf{X}_{\mathbf{a}}), \label{eq:best_traj}
    \end{align}
\end{subequations}
where \(\mathcal{X}_a\) represents a set of trajectory proposals optimized over the objective action \(\mathcal{J}_\mathbf{a}\). Each proposal is a time sequence of the SE(2) poses of the ego vehicle \(\mathbf{X}_{\mathbf{a}} = \left[ \left(\mathbf{p}^{\text{veh}}_{0,\tau}, \psi^{\text{veh}}_{0, \tau}\right) \right]^{t+T}_{\tau=t} \), with \(T\) standing for the motion planning horizon. 

Finally, we employ a trajectory selection mechanism that synthesizes high-level decision probability and low-level planning quality to arbitrate the best driving strategy from the multimodal trajectory candidate set \(\{\mathbf{X}^*_\mathbf{a}\}_{\mathbf{a} \in \mathcal{C}_t}\):
\begin{subequations}
    \begin{align}
    \mathbf{a}^{*} &= \arg\max_{\mathbf{a} \in \mathcal{C}_t} \Big[ \big( \hat{\pi}(\mathbf{a} \mid \xi_t) \big)^{\omega_{\text{prob}}} \cdot \widetilde{\mathcal{J}}_\mathbf{a}(\mathbf{X}^*_{\mathbf{a}}) \Big], \label{eq:final_action} \\
    \mathbf{X}^{\dagger} &= \mathbf{X}^*_{a^{*}},
    \end{align}\label{eq: final score}
\end{subequations}
where $\widetilde{\mathcal{J}}_{\mathbf{a}} = (\mathcal{J}_{\mathbf{a}}^{\text{dec}})^{\tilde{\omega}_{\text{dec}}} \cdot (\mathcal{J}^{\text{gen}})^{\tilde{\omega}_{\text{gen}}}$, with $\tilde{\omega}_{\text{dec}}, \tilde{\omega}_{\text{gen}}$ denoting adjusted balancing parameters during final arbitration. \(\omega_{\text{prob}} \in \mathbb{R}_+\) is the component for modulating the effect of decision uncertainty.

\subsection{Reflection System for Lifelong Learning}\label{sec:sys reflect}

\begin{figure*}[t]
  \centering
   \includegraphics[width=0.93\linewidth]{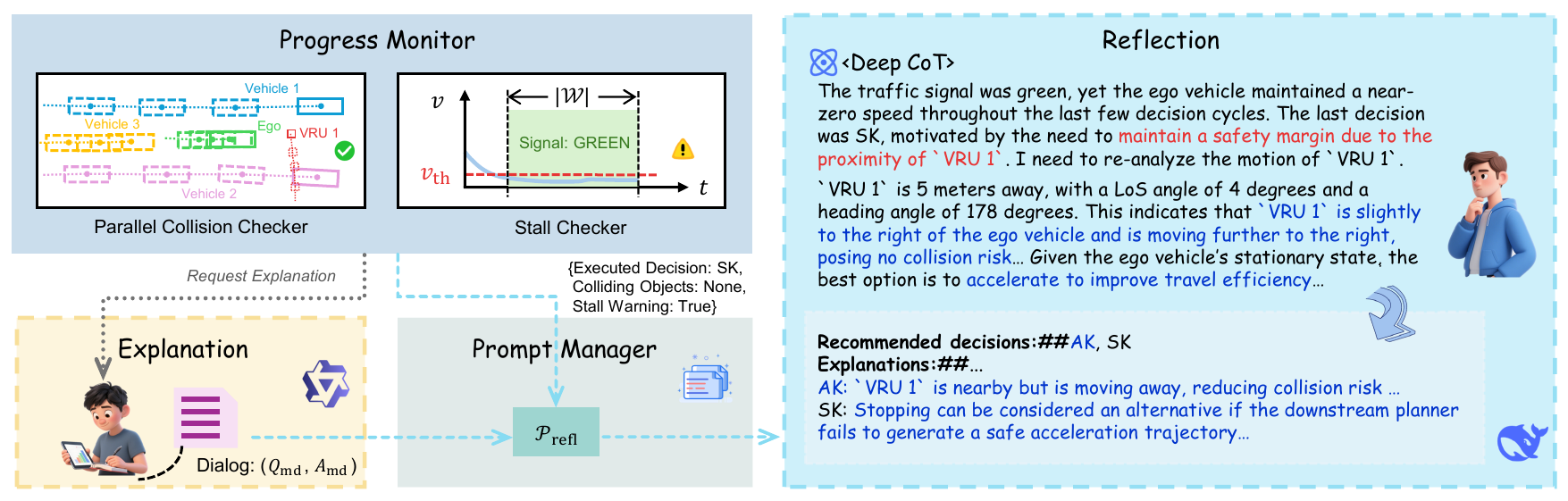}
   \caption{An illustration of the reflection mechanism. Upon the execution of a high-level decision in the closed-loop environment, the progress monitor performs system failure checking. If a collision happens or there is a possibility of a stall, the framework requests the heuristic system to generate an explanation of its previous decision-making process. The dialog of this decision-making, along with the system failure information, is sent to the prompt manager, which generates a query for the reflection system to analyze the situation and make possible revisions.}
   \label{fig:reflection}
\end{figure*}

Though with the well-established collaborative reasoning empowered by large foundation models, the analytical system may still produce unrealistic decisions in corner cases. Consequently, the heuristic system learning from these demonstrations may cause failures in similar closed-loop situations. We thus introduce a reflection system, paired with a progress monitor and an event-triggered decision explanation mechanism, to enable lifelong learning from failures and continuously update the framework's knowledge and adaptability.

\subsubsection{Failure Capture and Decision Explanation}\label{sec:failure explain}
We denote the time cycle of decision-making by \(T_{d}\). The progress monitor functions periodically according to \(T_{d}\) to examine driving progress, capturing potential system failures incurred by the previous decision. A potential failure triggers the heuristic system to generate a decision explanation.

\textbf{Failure Capture for Collisions and Stalls.} As shown in Fig.~\ref{fig:reflection}, two types of system failures, including collisions and stalls, are identified within the progress monitor. 

Let \(t\) denote the current decision-making time-step, the collision checker is deployed on a GPU device to calculate the overlap between the ego vehicle and surrounding objects during the last decision cycle \(\tau \in [t - T_d, t- T_d +1, \dots, t-1]\) in parallel. Specifically, we approximate road users and static objects as bounding boxes and implement the collision checking algorithm based on the Separating Axis Theorem (SAT). The pseudo code of the GPU-accelerated collision checker is shown in~\Cref{alg:collision}, which returns the index set of colliding objects $\mathcal{I}_{\text{coll}}$. 

For stall detection, we maintain a sliding window of the most recent $H$ decision cycles to monitor the ego vehicle's motion under permissible traffic conditions. Let $\mathcal{W} = \{ t - h T_d \mid h = 0, 1, \dots, H-1 \}$ denote the set of time-steps within this window. A warning $\Theta \in \{0,1\}$ is issued when the ego vehicle exhibits persistently low progress while traffic signals permit forward movement:
\begin{equation}\small
\Theta = \mathds{1}\left[ 
    \frac{1}{H} \sum_{\tau \in \mathcal{W}} v^{\text{veh}}_{0,\tau} < v_{\text{th}} 
    \;\land\;
    \bigwedge_{\tau \in \mathcal{W}} \bigl( \lambda_{\tau} \in \{\text{Green}, \text{None}\} \bigr)
\right],
\label{eq:stall}
\end{equation}
where $v_{\text{th}}$ is a velocity threshold, and $\lambda_{\tau}$ follows the traffic light phase definition in Table~\ref{tab:traffic states}. This condition captures scenarios where the vehicle remains quasi-stationary despite the absence of signal-mandated stopping requirements, indicating potential failure in making driving progress.

Finally, the tuple of system failures captured at the current time-step, \(\zeta_{t} \in \mathcal{S}_{\text{fail}}\) is constructed as:
\begin{equation}
    \zeta_{t} = \langle \tilde{\mathbf{a}}, \mathcal{I}_{\text{coll}}, \Theta \rangle,
\end{equation}
where \(\tilde{\mathbf{a}}\) denotes the decision issued at \(t-T_d\) and being executed till the current time-step.

\begin{algorithm}[t]
\caption{GPU-Accelerated Collision Checking}
\label{alg:collision}
\begin{algorithmic}[1]
\Require Bounding-box trajectories $\mathcal{T} \in \mathbb{R}^{N_{\text{obj}} \times T_d \times 4 \times 2}$, with ego vehicle at index $0$
\Ensure Collision indices $\mathcal{I}_{\text{coll}} \subseteq \{1,\dots,N_{\text{obj}}-1\}$

\If{$N_{\text{obj}} = 1$}
    \State \Return $\emptyset$
\EndIf

\State $\mathbf{V}^0 \gets \mathcal{T}[0]$ \Comment{Ego trajectory $(T_d,4,2)$}
\State $\mathbf{V} \gets \mathcal{T}[1:]$ \Comment{Others' trajectories $(N_{\text{obj}}-1,T_d,4,2)$}

\State \textit{// 1. Batch Dimension Alignment}
\State $\mathbf{V}^0_{\text{exp}} \gets \text{expand}(\mathbf{V}^0, (N_{\text{obj}}-1, T_d, 4, 2))$
\State $B \gets (N_{\text{obj}}-1) \times T_d$
\State $\mathbf{V}^0_{\text{flat}} \gets \text{reshape}(\mathbf{V}^0_{\text{exp}}, (B, 4, 2))$
\State $\mathbf{V}_{\text{flat}} \gets \text{reshape}(\mathbf{V}, (B, 4, 2))$

\State \textit{// 2. SAT Axis Construction}
\State $\mathbf{d}^0 \gets \text{roll}(\mathbf{V}^0_{\text{flat}}, -1, 1) - \mathbf{V}^0_{\text{flat}}$
\State $\mathbf{n}^0 \gets [-\mathbf{d}^0_{:, :, 1},\; \mathbf{d}^0_{:, :, 0}]^\top$ \Comment{Normals for ego}
\State $\mathbf{d} \gets \text{roll}(\mathbf{V}_{\text{flat}}, -1, 1) - \mathbf{V}_{\text{flat}}$
\State $\mathbf{n} \gets [-\mathbf{d}_{:, :, 1},\; \mathbf{d}_{:, :, 0}]^\top$ \Comment{Normals for others}
\State $\mathbf{N}_{\text{all}} \gets \text{concat}([\mathbf{n}^0, \mathbf{n}], \text{dim}=1)$ \Comment{$(B,8,2)$ axes}

\State \textit{// 3. Projection and Separation Test}
\State $\mathbf{P}^0 \gets \text{einsum}(\texttt{'bvk,bak->bav'}, \mathbf{V}^0_{\text{flat}}, \mathbf{N}_{\text{all}})$
\State $\mathbf{P} \gets \text{einsum}(\texttt{'bvk,bak->bav'}, \mathbf{V}_{\text{flat}}, \mathbf{N}_{\text{all}})$
\State $\text{min}^0, \text{max}^0 \gets \min(\mathbf{P}^0,2), \max(\mathbf{P}^0,2)$
\State $\text{min}, \text{max} \gets \min(\mathbf{P},2), \max(\mathbf{P},2)$
\State $\text{sep} \gets (\text{max}^0 < \text{min}) \lor (\text{max} < \text{min}^0)$ \Comment{$(B,8)$}
\State $\delta_{\text{inst}} \gets \neg \bigvee_{a=1}^{8} \text{sep}_{:,a}$ \Comment{Overlap if no separating axis}

\State \textit{// 4. Result Aggregation}
\State $\mathbf{M} \gets \text{reshape}(\delta_{\text{inst}}, (N_{\text{obj}}-1, T_d))$
\State $\delta \gets \bigvee_{\tau=0}^{T_d-1} \mathbf{M}_{:, \tau}$ \Comment{Collision if overlap at any timestep}
\State $\mathcal{I}_{\text{coll}} \gets \{ i+1 \mid \delta_{i} = \text{True} \}$
\State \Return $\mathcal{I}_{\text{coll}}$
\end{algorithmic}
\end{algorithm}

\textbf{Failure-Triggered Decision Explanation.} Under the condition of potential system failures, \ie, \((\mathcal{I}_{\text{coll}} \neq \emptyset ) \vee \Theta\), the framework triggers the heuristic system to autoregressively generate an explanation of the previous decision. The generation process is asynchronous to the periodic decision-making, without deteriorating real-time inference efficiency. To align the explanation with the predicted decision distribution, we first transform each action \(\mathbf{a} \in \mathcal{C}_t\) into its prescribed linguistic notation via \(\Xi: \mathcal{A} \to \mathcal{L}_{\text{nat}}\) (\eg, \(\text{Cruise} \times \text{Lane Keep} \mapsto \texttt{CK}\)). These notations are then sorted in descending order of their associated probabilities \(\hat{\pi}(\cdot \mid \xi_t)\) to form an ordered sequence, as required by the response structure in demonstrations (see the third row of \Cref{tab:agent-specific instruction}):
\begin{equation}\small
    \big( \Xi(\mathbf{a}_{(1)}), \dots, \Xi(\mathbf{a}_{(|\mathcal{C}_t|)}) \big) 
    = \text{argsort}_{\downarrow}\Big( \big\{ \hat{\pi}(\mathbf{a} \mid \xi_t) \big\}_{\mathbf{a} \in \mathcal{C}_t} \Big),
    \label{eq:sorted_notations}
\end{equation}
where \(\Xi(\mathbf{a}_{(1)}), \dots, \Xi(\mathbf{a}_{(|\mathcal{C}_t|)})\) denotes the probability-ranked action notations. The query conditioned on the predicted distribution is constructed by concatenating a description of these sorted actions to the augmented \(\bar{Q}_{\text{md}}\) query from Eq.~\eqref{eq:augmented_query}:
\begin{subequations}
    \begin{align}
        \Delta A_{\text{md}} &= u_{\text{pref}} \;\Vert\; \big( \Xi(\mathbf{a}_{(1)}) \Vert u_{\text{sep}} \Vert \cdots \Vert \Xi(\mathbf{a}_{(|\mathcal{C}_t|)}) \big), \\
        \bar{Q}^{+}_{\text{md}} &= \bar{Q}_{\text{md}} \;\Vert\; \Delta A_{\text{md}},
        \label{eq:failure_augmented_query}
    \end{align}
\end{subequations}
where \(\Delta \hat{A}_{\text{md}}\) stands for the description of sorted actions, with \(u_{\text{pref}}\) representing the explanation prefix (\ie, ``\(\texttt{Recommended decisions:\#\#}\)") and \(u_{\text{sep}}\) denoting the separator (\ie, ``\(\texttt{,}\)"). The language model then autoregressively generates the subsequent explanation content:
\begin{equation}
    A^{-}_{\text{md}} = \mathcal{G}_{6}\big( \bar{Q}^{+}_{\text{md}}; \underline{\tau}, \theta \big),
    \label{eq:failure_explanation}
\end{equation}
where \( \mathcal{G}_{6}\big( \, \cdot \, ; \underline{\tau} , \theta \big)\) denotes the autoregressive generation process implemented by the fine-tuned language model \(f_{\theta}\) (Sec.~\ref{sec:sys heuristic}) under zero temperature. The complete decision explanation content is recovered through concatenation as:
\begin{equation}
    A_{\text{md}} = \Delta A_{\text{md}} \; \Vert \; A^{-}_{\text{md}}.
\end{equation}

\subsubsection{Reflection Process and Memory Update}
Based on the dialog of multimodal decision-making and system failure information, the prompt manager generates a query to the reflection system, initiating the reflection process:
\begin{subequations}
    \begin{align}
        Q_{\text{refl}} &= \mathcal{P}_{\text{refl}} \Big(Q_{\text{md}}, A_{\text{md}}, \zeta_t, \mathbf{K}_{\text{refl}}(\mathcal{B},\mathcal{A}) \Big), \\
        A_{\text{refl}} &= \mathcal{G}_7(Q_{\text{refl}};\underline{\tau}),
    \end{align}
\end{subequations}
where \(\mathcal{G}_7\) represents the text generation process of reflection. Herein, we use the original query \(Q_{\text{md}}\) without retrieval augmented context to form the dialog of decision-making, in order to have the reflection system focus precisely on the driving situation in which potential failures are captured. The key portion of the reflection instruction \(\mathbf{K}_{\text{refl}}(\mathcal{B},\mathcal{A})\) is displayed in the last row of \Cref{tab:agent-specific instruction}.

The reflection system is allowed to make two types of responses: a revised decision explanation or a sign of remaining unchanged, where the latter can only be issued when the situation does not involve collisions and the decision for conservative driving is well justified. Upon receiving a revised decision explanation, the framework accumulates the item \(\upsilon(\xi_{t-T_d};\zeta_t) = \langle Q_{\text{md}}, \mathbf{e}, -, A_{\text{refl}}\rangle\) to the memory bank to complete the update. The memory items obtained through reflection do not possess labels of decision distributions. Hence, these items are only utilized in RAG to advance future driving tasks.

\section{Experiment Setup}
\subsection{Evaluation Methods and Datasets}
We systematically evaluate our approach in both closed-loop and open-loop settings, using datasets including nuPlan~\cite{caesar2021nuplan}, DriveLM-nuScenes~\cite{sima2024drivelm}, and our in-house dataset collected through field experiments.

\subsubsection{Closed-Loop Experiments}
LUNA-AD is tested on the nuPlan \textbf{Test14-Hard} and \textbf{Test14-Random} benchmarks~\cite{cheng2024rethinking}, which comprise 14 scenario types, including long-tail situations in the Hard set. Experiments run in \textbf{Non-Reactive} (\textbf{NR}) mode (log-replay traffic) and \textbf{Reactive} (\textbf{R}) mode (Intelligent-Driver-Model (IDM)~\cite{helbing1998generalized} traffic). We use standard nuPlan metrics for evaluation: the Closed-Loop Scores (\textbf{NR-CLS}/\textbf{R-CLS}) aggregate performance on multiple aspects, including efficiency, comfort, and human-likeness, while the Success Rates (\textbf{NR-SR(\%)}/\textbf{R-SR(\%)}) quantify the avoidance of severe errors. Specifically, Success Rate is the proportion of runs not receiving a \textit{zero} score due to collisions, off-road excursions, incorrect direction, or low progress. In the simulation stack, a linear quadratic regulator tracks the trajectory to produce control inputs for a kinematic bicycle model, which handles the vehicle dynamics.

Our approach is comparatively evaluated against four classes of methods: \textbf{rule-based}, \textbf{data-driven}, \textbf{hybrid (rule+data)}, and \textbf{knowledge-driven methods}.
For fair comparison, we benchmark the open-source implementation of different methods on a four-card RTX-4090 computing device in our evaluation. For methods remaining closed-source, the results are quoted from the original publications.

\subsubsection{Open-Loop Experiments}
\begin{figure}[t]
  \centering
   \includegraphics[width=0.9\columnwidth]{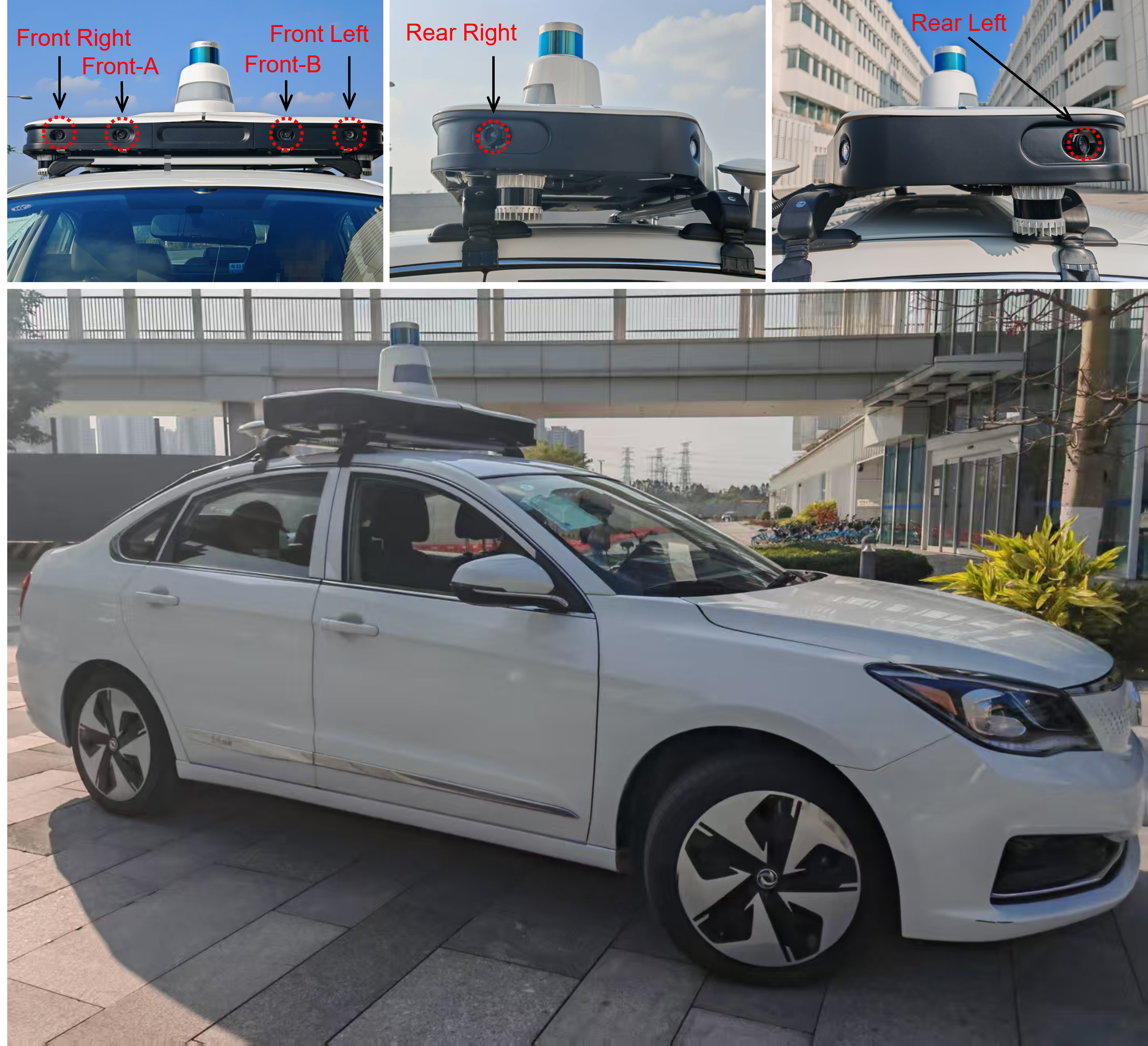}
   \caption{The camera system of the data collection vehicle. Six RGB cameras are deployed: two front (Front-A, Front-B), front-left, front-right, rear-left, and rear-right, covering full surrounding scenes.}
   \label{fig:field exp vehicle}
\end{figure}

The open-loop experiments include: \textbf{few-shot effects on the heuristic system} and \textbf{domain transfer with vision inputs}.

First, we test the fine-tuned heuristic system on decision prediction and language generation under varied few-shot settings. Using 2K nuPlan frames (excluded from SFT), we evaluate decision prediction via accuracy (matching highest-probability decisions) and KL divergence (Eq.~\eqref{eq:dpp_loss}). Language generation is assessed with BLEU-1, BLEU-4, METEOR, and ROUGE-L.

Second, we evaluate transferability of our multi-agent analytical system to RGB inputs. A variant with a different set of prompts while retaining the fundamental structure of action voting, confidence assessment, and summarization is tested on 300 DriveLM-NuScenes frames. It processes six-view RGB images (front, front-right, front-left, back, back-right, back-left). The action space is modified to \(\mathcal{A}^{\dagger}=\{\text{Fast}, \text{Normal Speed}, \text{Slow}, \text{Stop}\} \times \{\text{Left}, \text{Right}, \text{Straight}\}\), merging fine-grained categories due to VLM limitations~\cite{xie2025vlms}. We report Accuracy-\(K\): frequency that ground-truth falls within top-\(K\) predictions.

Finally, a field experiment is conducted on a college campus using an intelligent vehicle (camera system in Fig.~\ref{fig:field exp vehicle}, 10\,Hz RGB). Each sample contains 2-second multi-view video history at 2\,Hz, unlike DriveLM-NuScenes single-frame inputs. We deploy our analytical system to test performance beyond standard urban scenarios.

\begin{table*}[ht]
    \centering
    \begin{minipage}{0.95\textwidth}
        \caption{Quantitative closed-loop performance of different approaches. All metrics are \underline{higher the better}. The best performance is highlighted in \colorbox{lightblue}{lightblue} and the second best performance highlighted in \colorbox{newlightgreen}{olive green}.}
        \label{tab:planner compare}
    \end{minipage}
    
    \vspace{-0.2cm} 

    \fontsize{7}{8.4}\selectfont 
    \begin{threeparttable}
    \begin{tabular}{@{}p{0.22\columnwidth}p{0.28\columnwidth}M{0.12\columnwidth}M{0.135\columnwidth}M{0.12\columnwidth}M{0.12\columnwidth}M{0.12\columnwidth}M{0.14\columnwidth}M{0.12\columnwidth}M{0.12\columnwidth}@{}}
        \toprule
        \multirow{2}{*}{\textbf{Type}} & \multirow{2}{*}{\textbf{Model}} & \multicolumn{4}{c}{\textbf{Test14-Hard}} & \multicolumn{4}{c}{\textbf{Test14-Random}} \\ 
        \cmidrule(lr){3-6} \cmidrule(lr){7-10}
        & & \textbf{NR-CLS}  & \textbf{NR-SR(\%)}  & \textbf{R-CLS}  & \textbf{R-SR(\%)}  & \textbf{NR-CLS}  & \textbf{NR-SR(\%)}  & \textbf{R-CLS}  & \textbf{R-SR(\%)} \\ 
        \midrule
        \textcolor{mediumgray}{Expert} & \textcolor{mediumgray}{Log-Replay} & \textcolor{mediumgray}{85.96} & \textcolor{mediumgray}{91.18} & \textcolor{mediumgray}{68.80} & \textcolor{mediumgray}{73.53} & \textcolor{mediumgray}{94.03} & \textcolor{mediumgray}{96.93} & \textcolor{mediumgray}{75.86} & \textcolor{mediumgray}{78.16}\\
        \midrule
        \multirow{2}{*}{Rule-Based} & IDM~\cite{helbing1998generalized} & 56.16 & 65.81 & 62.26 & 71.69 & 70.39 & 77.01 & 72.42 & 78.54 \\ 
        & PDM-Closed\(^{\dag}\)~\cite{dauner2023parting} & 65.08 & 78.68 & 75.19 & 85.66 & 90.05 & 94.64 & \makecell{\colorbox{lightblue}{91.64}} & \makecell{\colorbox{lightblue}{96.55}} \\ 
        \midrule
        \multirow{5}{*}{Data-Driven} & GC-PGP~\cite{hallgarten2023prediction} & 47.51 & 56.62 & 42.80 & 51.84 & 61.95 & 72.03 & 56.46 & 65.13 \\
                                                 & UrbanDriverOL~\cite{scheel2022urban} & 51.67 & 59.20 & 49.06 & 55.51 & 63.57 & 70.88 & 60.92 & 66.28 \\
                                                 & RasterModel~\cite{caesar2021nuplan} & 50.65 & 58.09 & 52.44 & 59.56 & 67.70 & 73.56 & 68.64 & 74.33 \\
                                                 & PlanTF~\cite{cheng2024rethinking} & 72.56 & 79.04 & 60.34 & 67.28 & 85.60 & 90.80 & 78.86 & 85.44 \\
                                                 & Diffusion-Planner\cite{zhengdiffusion} & 75.29 & 82.72 & 68.83 & 78.31 & 88.98 & 93.87 & 83.21 & 90.42 \\
        \midrule
        \multirow{4}{*}{Hybrid} & PDM-Hybrid\(^{\dag}\)~\cite{dauner2023parting} & 65.99 & 79.41 & 76.07 & 86.40 & \makecell{\colorbox{newlightgreen}{90.10}} & 95.02 & \makecell{\colorbox{newlightgreen}{91.29}} & \makecell{\colorbox{newlightgreen}{96.17}} \\
                                         & GameFormer~\cite{huang2023gameformer} & 62.94 & 74.26 & 61.53 & 72.79 & 76.89 & 85.29 & 77.80 & 86.59 \\
                                         & Diffusion-ES~\cite{yang2024diffusion} & \makecell{\colorbox{newlightgreen}{77.54}} & 84.93 & \makecell{\colorbox{newlightgreen}{77.75}} & 87.13 & 87.70 & 94.25 & 87.18 & 93.87 \\
                                         & PLUTO\(^{\dag}\)~\cite{cheng2024pluto} & \makecell{\colorbox{lightblue}{79.19}} & 88.24 & 75.75 & 84.19 & \makecell{\colorbox{lightblue}{91.93}} & \makecell{\colorbox{newlightgreen}{95.79}} & 89.94 & \makecell{\colorbox{newlightgreen}{96.17}} \\
        \midrule
        \multirow{4}{*}{Knowledge-Driven} & PlanAgent\(^*\)~\cite{zheng2024planagent} & 72.51 & - & 76.82 & - & - & - & - & - \\
                                                & CALMM-Drive-L~\cite{yao2024calmm}\(^\ddagger\) & 77.39 & \makecell{\colorbox{newlightgreen}{88.97}} & \makecell{\colorbox{lightblue}{78.13}} & \makecell{\colorbox{newlightgreen}{89.71}} & 86.70 & 95.02 & 87.11 & 95.79 \\
                                                & CALMM-Drive-M~\cite{yao2024calmm}\(^\ddagger\) & 70.04 & 84.19 & 67.85 & 80.88 & 81.82 & 92.72 & 83.37 & 95.02 \\
                                       & LUNA-AD (ours) & 76.11 & \makecell{\colorbox{lightblue}{89.71}} & 76.78 & \makecell{\colorbox{lightblue}{90.44}} & 86.98 & \makecell{\colorbox{lightblue}{96.93}} & 87.02 & \makecell{\colorbox{lightblue}{96.55}}\\
        \bottomrule
    \end{tabular}
    \vspace{0.1cm} 

    \begin{tablenotes}
    \item[$\dagger$] Methods with rule-based modules for emergency braking. 
    \item[$*$] Quoted from original publications.
    \item[$\ddagger$] CALMM-Drive-L and CALMM-Drive-M are instantiated by GPT-4o and GPT-4o-mini respectively.
    \end{tablenotes}
    
    \end{threeparttable}
    \vspace{-0.4cm}
\end{table*}

\begin{table}[t]
\centering

\begin{minipage}{\columnwidth}
    \caption{Inference efficiency of LLM-based decision-making. 
    Latency for API-based methods includes network transmission.}
    \label{tab:efficiency}
\end{minipage}

\vspace{-0.2cm} 

\fontsize{7}{8.4}\selectfont 
\begin{threeparttable}
\begin{tabular}{p{0.28\columnwidth}M{0.17\columnwidth}M{0.17\columnwidth}M{0.17\columnwidth}}
\toprule
\textbf{Model} & \textbf{Parameters} & \textbf{Latency} & \textbf{Peak VRAM}\\
\midrule
PlanAgent~\cite{zheng2024planagent} & $\sim$\,280\,B$^\dagger$ & 5568\,ms$^*$ & N/A$^\ddagger$ \\
CALMM-Drive-L~\cite{yao2024calmm} & $\sim$\,200\,B$^\dagger$ & 4747\,ms & N/A$^\ddagger$\\
CALMM-Drive-M~\cite{yao2024calmm} & $\sim$\,8\,B$^\dagger$ & 2982\,ms & N/A$^\ddagger$\\
LUNA-AD (Ours) & \makecell{\colorbox{lightblue}{1.7\,B}} & \makecell{\colorbox{lightblue}{192\,ms}} & 3671\,MB\\
\bottomrule
\end{tabular}

\vspace{0.1cm}

\begin{tablenotes}
\item[$\dagger$] Closed-source foundation models; estimates indicating activated parameters based on widely recognized third-party reverse-engineering analyses (\eg, SemiAnalysis), reported solely for relative scale comparison.
\item[$*$] Quoted from original publications.
\item[$\ddagger$] VRAM required for API-based methods are not available.
\end{tablenotes}
\end{threeparttable}
\vspace{-0.4cm}
\end{table}

\subsection{Implementation Details}
\textbf{Multi-Agent Analytical System.} Our primary setup (evaluated in \Cref{sec:closed loop,sec:effects fewshot}) uses DeepSeek-V3~\cite{liu2024deepseek} for action voting and confidence assessment, with temperature \(\widetilde{\tau}=0.7\) and sample size \(M=10\) to promote diversity. DeepSeek-R1~\cite{guo2025deepseek} handles summarization for its deep reasoning on long contexts. The system generates 10.2\,K memory entries from nuPlan training data at 2\,Hz.

For the vision-based variant (\Cref{sec:domain transfer}), GPT-4o implements voting/assessment, Gemini-3-Pro does summarization, with \(\widetilde{\tau}\in\{0.3,0.5,0.7,0.9,1.1\}\) and \(M=10\).

\textbf{Multi-Task Heuristic System.} Qwen3-1.7B~\cite{yang2025qwen3} is the base LM. The classification head is a two-layer MLP (hidden dim 1024). We use 4-bit quantization + LoRA (r=16), with LoRA adapters and classifier in bfloat16, modifying only 0.08\% of parameters. Training uses gradient checkpointing and 8-bit AdamW on four RTX-4090 GPUs for 20 epochs ($\sim$35 hours). Hyperparameters: \(\alpha = 0.7\), \(K_{\text{min}} = 0\), \(K_{\text{max}} = 3\). Inference uses 3 few-shot examples.

\textbf{Reflection System.} DeepSeek-R1 serves as the foundation model for its deep reasoning and fine-grained mistake capture. We deploy the framework in nuPlan Reduced-Val14~\cite{dauner2023parting} (318 scenarios) to detect failures, perform reflections, and update the memory bank before testing on Test14-Random and Test14-Hard.

\textbf{Decision-Guided Planner.} Candidate actions use confidence threshold \(\gamma_p = 0.1\). Scoring weights: \(\omega_{\text{dec}} = 5.0\), \(\omega_{\text{gen}} = 1.0\), \(\omega_{\text{prob}} = 1.0\), \(\tilde\omega_{\text{dec}} = 0.1\), \(\tilde\omega_{\text{gen}} = 0.3\). The planner discretizes motion at 0.1\,s intervals over a 40-step horizon. Other hyperparameters follow~\cite{yao2024calmm}.

\section{Results and Analysis}

\begin{table*}[ht]
    \centering
    \begin{minipage}{\textwidth}
        \caption{Results of ablation study on key components in the overall architecture.}
        \label{tab:ablation}
    \end{minipage}

    \vspace{-0.2cm}
    
    \resizebox{\textwidth}{!}{
    \begin{threeparttable}
    \begin{tabular}{@{}p{2.3cm}cccccc|cccccccc@{}}
        \toprule
        \multirow{2}{*}{\textbf{Model}} & 
        \multicolumn{6}{c|}{\textbf{Configuration}} & 
        \multicolumn{4}{c}{\textbf{Test14-Hard}} & 
        \multicolumn{4}{c}{\textbf{Test14-Random}} \\ 
        \cmidrule(lr){2-7} \cmidrule(lr){8-11} \cmidrule(lr){12-15}
        & \textbf{Mem.} & \textbf{A.V.} & \textbf{C.A.} & \textbf{U.A.} & \textbf{Sum.} & \textbf{Refl.} & 
        \textbf{NR-CLS} & \textbf{NR-SR(\%)} & \textbf{R-CLS} & \textbf{R-SR(\%)} & 
        \textbf{NR-CLS} & \textbf{NR-SR(\%)} & \textbf{R-CLS} & \textbf{R-SR(\%)} \\
        \midrule
        $\mathcal{M}_0$ & \checkmark & $\times$ & $\times$ & $\times$ & $\times$ & $\times$ & 73.91 & 85.66 & 72.53 & 85.66 & 84.56 & 93.87 & 83.81 & 94.64 \\
        $\mathcal{M}_1$ & \checkmark & \checkmark & $\times$ & $\times$ & $\times$ & $\times$ & 74.38 & 87.13 & 73.39 & 86.76 & 85.53 & 95.79 & 85.13 & 95.40 \\
        $\mathcal{M}_2$ & \checkmark & \checkmark & \checkmark & $\times$ & $\times$ & $\times$ & 74.37 & 86.76 & 74.98 & 88.24 & 86.01 & 96.55 & 85.33 & 95.40 \\
        $\mathcal{M}_3$ & \checkmark & \checkmark & \checkmark & \checkmark & $\times$ & $\times$ & 74.90 & 87.50 & 75.50 & 88.60 & 85.56 & 95.40 & 86.31 & 96.17 \\
        $\mathcal{M}_4$ & \checkmark & \checkmark & \checkmark & \checkmark & \checkmark & $\times$  & \makecell{\colorbox{lightblue}{76.26}} & 89.34 & 76.25 & 89.71 & 86.41 & \makecell{\colorbox{lightblue}{96.93}} & 86.37 & 96.17 \\
        \midrule
        LUNA-AD & \checkmark & \checkmark & \checkmark & \checkmark & \checkmark & \checkmark & 76.11 & \makecell{\colorbox{lightblue}{89.71}} & \makecell{\colorbox{lightblue}{76.78}} & \makecell{\colorbox{lightblue}{90.44}} & \makecell{\colorbox{lightblue}{86.98}} & \makecell{\colorbox{lightblue}{96.93}} & \makecell{\colorbox{lightblue}{87.02}} & \makecell{\colorbox{lightblue}{96.55}} \\
        LUNA-AD (0-shot) & $\times$ & \checkmark & \checkmark & \checkmark & \checkmark & \checkmark & 73.83 & 88.24 & 73.85 & 87.50 & 85.93 & 95.40 & 82.39 & 92.34\\
        \bottomrule
    \end{tabular}%
    \end{threeparttable}
    }
\vspace{-0.3cm}
\end{table*}

\subsection{Closed-Loop Evaluations}\label{sec:closed loop}

\subsubsection{Comparative Studies}\label{sec:compare results}
\Cref{tab:planner compare} presents closed-loop evaluations across four method categories under NR and R modes. Key observations are summarized below.

\textbf{LUNA-AD achieves SOTA success rates across all settings.} On Test14-Hard, it attains 89.71\,\% NR-SR and 90.44\,\% R-SR, surpassing all competitors. On Test14-Random, it reaches 96.93\,\% NR-SR (matching expert log-replay) and 96.55\,\% R-SR, demonstrating robust handling of both standard and distribution-shifted scenarios.

\textbf{Our method shows substantially narrower performance degradation from common to long-tail scenarios.} The NR-SR drop from Test14-Random to Test14-Hard is only 7.22\,\%, versus 11.15\,\% for Diffusion-Planner and 9.32\,\% for Diffusion-ES. Among knowledge-driven methods, it narrows the gap relative to CALMM-Drive-M (8.53\,\%) while matching CALMM-Drive-L's robustness (6.05\,\%).

\textbf{Our approach maintains exceptional stability under reactive traffic.} When handling interactive vehicles, LUNA-AD's SR changes by only +0.73\,\% (Test14-Hard) and -0.38\,\% (Test14-Random), sharply contrasting with PLUTO's 4.05 point drop and PlanTF's 11.76 point degradation on Test14-Hard. This shows our probabilistic representation safely accommodates emergent interactions.

\textbf{LUNA-AD achieves competitive closed-loop scores while securing the highest success rates.} On Test14-Hard, it obtains 76.11 NR-CLS and 76.78 R-CLS, remaining 3.08 points and 1.35 points behind the SOTA methods. Critically, this CLS accompanies the highest SR, reflecting a safety-oriented trade-off where SR penalizes catastrophic failures while CLS aggregates softer metrics.

\Cref{tab:efficiency} compares the inference efficiency of LLM-based decision-making modules. \textbf{LUNA-AD's average latency (192\,ms) is over an order of magnitude lower than other methods,} thanks to its lightweight architecture and test-time disentanglement of deliberative reasoning. \textbf{Its peak VRAM consumption (3671\,MB) enables deployment on a single consumer-grade GPU.} Collectively, these comparisons validate LUNA-AD's superiority in both closed-loop performance and practical deployability.

\subsubsection{Ablation Studies}\label{sec:ablation}
We evaluate several ablated versions of our approach:

\begin{itemize}
    \item \(\mathcal{M}_0\): Baseline learning from a unimodal decision-making teacher (\(\widetilde\tau=0\)) without multi-agent action voting (A.V.).
    \item \(\mathcal{M}_1\): Adds A.V. but omits confidence assessment (C.A.), using normalized frequencies as one-hot labels and arbitrary top-category language responses.
    \item \(\mathcal{M}_2\): Uses C.A. and Eq.~\eqref{eq:probabilistic_synthesis} but converts probabilities to one-hot labels, disabling uncertainty awareness (U.A.); it employs the highest-confidence response within the top category for language modeling.
    \item \(\mathcal{M}_3\): Applies multimodal supervision via Eq.~\eqref{eq:probabilistic_synthesis}, but excludes the summarization (Sum.) agent, retaining the text annotations of \(\mathcal{M}_2\).
    \item \(\mathcal{M}_4\): Combines the full multi-agent analytical and multi-task heuristic system, excluding the reflection system.
    \item LUNA-AD (0-shot): Full framework without RAG during testing.
\end{itemize}

As shown in~\Cref{tab:ablation}, the full LUNA-AD achieves superior performance on nearly all metrics, except for a 0.15-point decrease on Test14-Hard NR-CLS compared to \(\mathcal{M}_4\). The vanilla \(\mathcal{M}_0\) yields the lowest performance. A.V. in \(\mathcal{M}_1\) enables deeper exploration of the teacher's reasoning, uncovering robust solutions and improving closed-loop outcomes. \(\mathcal{M}_2\) leverages explicit confidence assessment to select higher-quality responses, surpassing \(\mathcal{M}_1\) on most criteria. \(\mathcal{M}_3\) adopts soft labels to capture decision uncertainty, promoting efficient distillation and boosting performance. \(\mathcal{M}_4\) includes a summarization agent to produce textual explanations grounded in multimodal decisions, enriching supervision alignment. Finally, the full LUNA-AD integrates a reflection mechanism to continually refresh its knowledge base, enhancing adaptation to closed-loop scenarios. Notably, LUNA-AD (0-shot) suffers significant performance drops in reactive modes, underscoring the importance of RAG for interaction-aware decision-making.

\subsubsection{Qualitative Studies}\label{subsec: quali}
\begin{figure*}
  \centering
    \includegraphics[
    width=\linewidth]{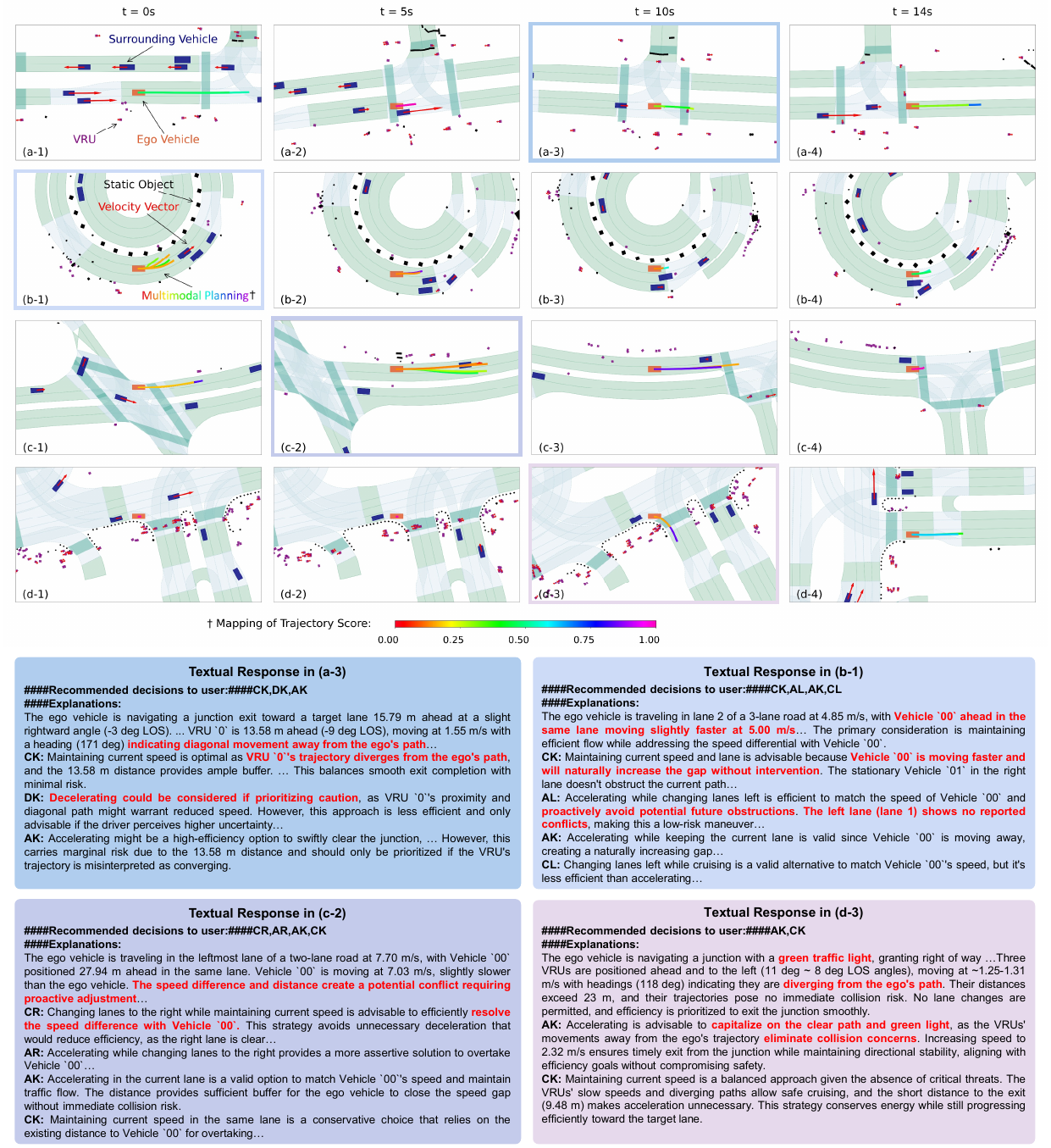}
  \caption{Qualitative demonstrations of our approach in four representative scenarios. For better visualization, the scores of different planning trajectories are normalized to sum to one after being computed using Eq.~\eqref{eq: final score}. In textual responses, we highlight the key contents in interaction analysis in \textcolor{red}{red}.}
  \label{fig:qualitative}
  \vspace{-0.5cm}
\end{figure*}

We showcase our framework's ability to generate multimodal driving strategies with explanations in four representative scenarios, as shown in Fig.~\ref{fig:qualitative}. The first two examples are generated without reflection-based memory update, while others adopt the updated memory bank.

\textbf{(a-1) - (a-4).} Starting on a multi-lane roadway, the ego vehicle approaches an intersection. At \(t=0\,\text{s}\), with clear conditions ahead, the top-ranked trajectory adopts an acceleration policy for efficient junction traversal. When a VRU is encountered at \(t=10\,\text{s}\), the system evaluates the VRU's trajectory; recognizing its departing motion, a cruise maneuver is selected to jointly optimize safety and efficiency. 

\textbf{(b-1) - (b-4).} Snapshot `b-1' depicts a multimodal planning process involving lane-change operations. Regarding high-level decisions, `CK' and `AL' are identified as the most viable options, where `CK' is favored owing to the leader vehicle's sustained higher speed. As the leader vehicle then decelerates to accommodate a VRU and a cut-in vehicle, the ego vehicle performs a lane change to the left, thus avoiding impediments and maintaining efficient progress.

\textbf{(c-1) - (c-4).} Our LUNA-AD initiates a multimodal planning process to address interaction with the leading vehicle in snapshot `c-2'. With the speed of the ego vehicle being higher than that of the leading vehicle, it adopts a lane-changing maneuver to proactively detour around. This action also avoids obstruction by two VRUs that move slowly on the road. Finally, the ego vehicle reaches a junction and remains in the correct lane to prepare for a right turn.

\textbf{(d-1) - (d-4).} The ego vehicle encounters a group of VRUs crossing the street and remains nearly stationary in the first few seconds. At \(t=10\,\text{s}\), our framework rules out collision risks with VRUs and confirms a green light, then adopts an acceleration maneuver to swiftly complete the right turn. Meanwhile, the multimodal decision-maker deems cruising an alternative given that cruising offers energy-saving benefits.

\begin{figure*}
  \centering
    \includegraphics[width=0.85\linewidth, trim=0 30 0 0, clip]{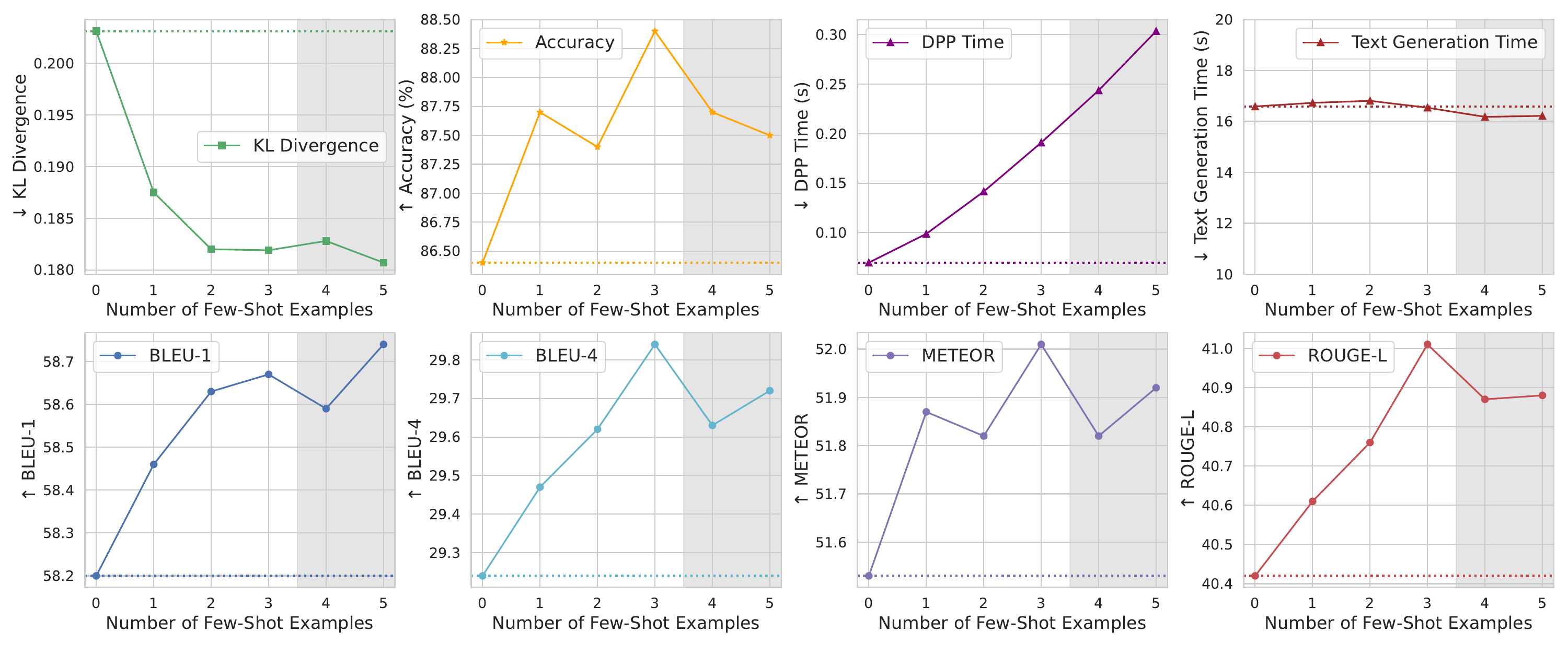}
  \caption{Effects of test-time RAG few-shot count on the open-loop performance of the heuristic system, which includes the fitness of decision probability prediction, the quality of language generation, and the inference latency of the two tasks. Shaded regions denote few-shot settings unseen during training.}
  \label{fig:param effects}
  \vspace{-0.5cm}
\end{figure*}

\subsection{Open-Loop Evaluations}

\subsubsection{Effects of the Number of Few-Shots}\label{sec:effects fewshot}
As shown in Fig.~\ref{fig:param effects},
\textbf{RAG improves prediction accuracy and calibration.} Compared to zero-shot, few-shot examples reduce KL divergence from 0.2031 to 0.1807 and increase accuracy from 86.40\,\% to 88.40\,\%. While the overall optimal performance is achieved with 3 examples (88.40\,\% accuracy, 0.1819 KLD), the performance with 4--5 examples (unseen during training) remains superior to zero-shot.
\textbf{The peaks of language generation quality vary by metric.} BLEU4 (29.84), METEOR (52.01), and ROUGE-L (41.01) peak at 3 examples, while BLEU1 peaks at 5 examples (58.74), showing different responses to context length. All metrics consistently exceed zero-shot performance.
\textbf{Decision inference time increases with few-shot count but stays below 0.2\,s for up to 3 examples}, achieving significant speedup over competitive knowledge-driven methods while preserving multimodal reasoning. Language generation time remains stable around 16.5 seconds. In closed-loop deployment, this task is event-triggered (\Cref{sec:failure explain}), ensuring low latency and interpretability.

\subsubsection{Domain Transfer Studies with Vision Input Modalities}\label{sec:domain transfer}

\begin{figure*}
  \centering
    \includegraphics[
    width=0.95\linewidth]{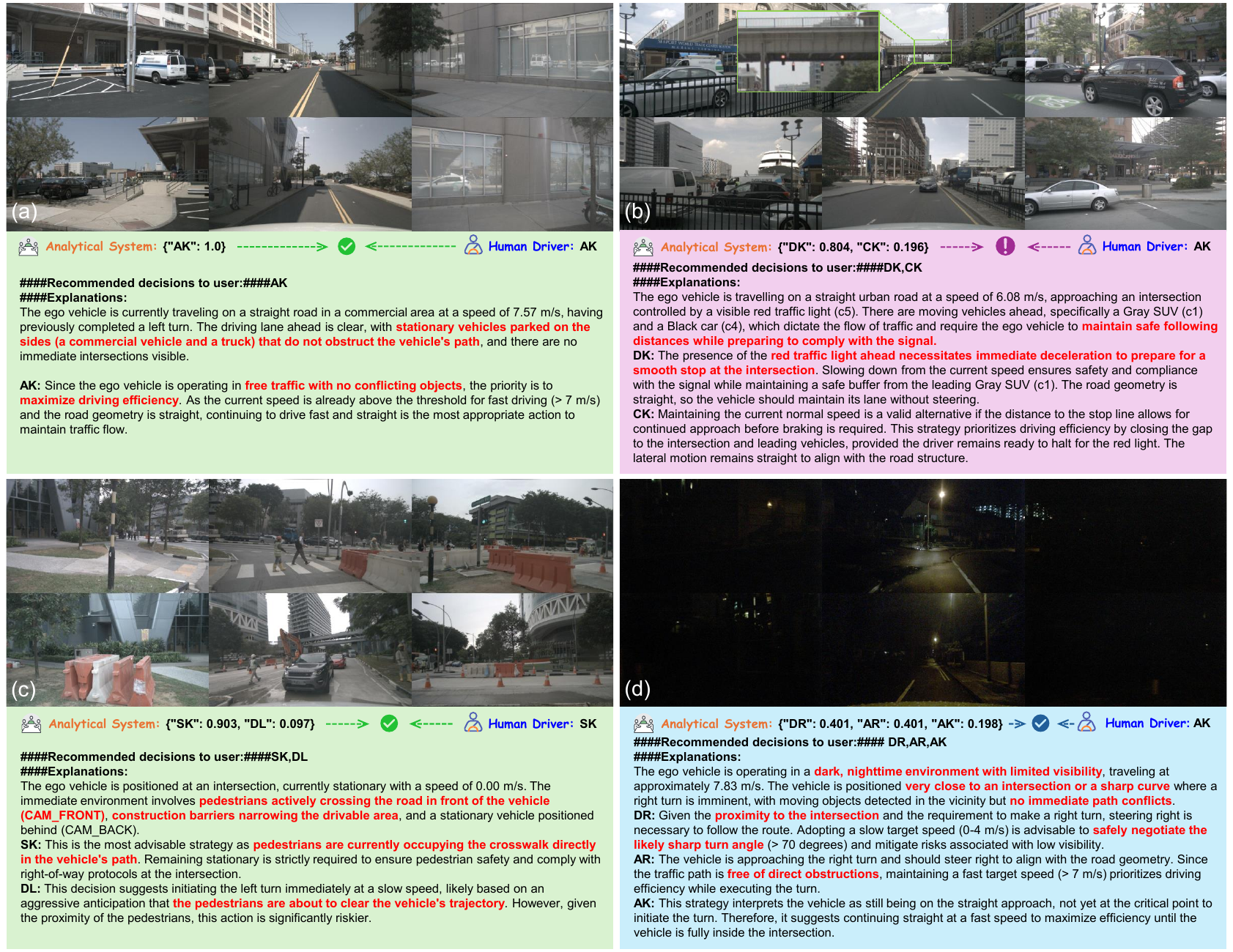}
  \caption{Representative cases from DriveLM-NuScenes for demonstrating the analytical system's ability to perform multimodal decision reasoning with multi-view image inputs. Symbols: \inlineicon{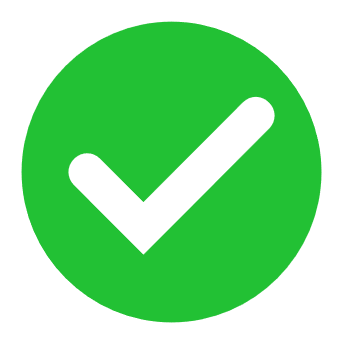}: align with human driver under Acc-1; \inlineicon{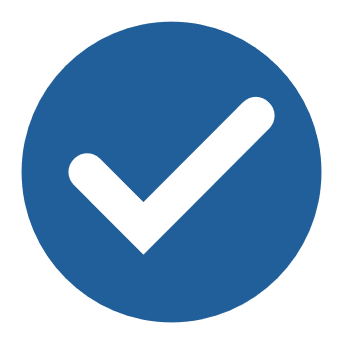}: align with human driver under Acc-2 or Acc-3; \inlineicon{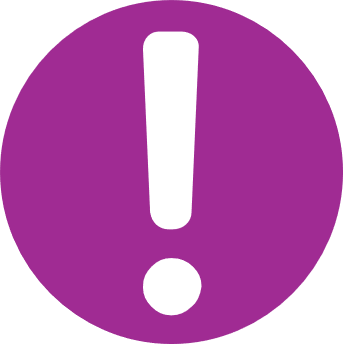}: not align with human driver. Action Codes: \{Fast: A, Normal Speed: C, Slow: D, Stop: S\} \(\times\) \{Left: L, Right: R, Straight: K\}. Key contents in responses are highlighted in \textcolor{red}{red}.}
  \label{fig:vis drivelm}
  \vspace{-0.5cm}
\end{figure*}

\textbf{Evaluations using DriveLM-NuScenes.} \Cref{tab:temperature scaling} reports our analytical system's performance under varying temperatures, compared with A.V.\,-\,Only (single agent, zero temperature). Acc-1 and Acc-3 peak at 65.00\,\% and 78.33\,\% when \(\widetilde{\tau}=0.7\), indicating that moderate stochasticity encourages multimodal strategies that better reflect human driving. Beyond 0.7, Acc-1 drops due to excessive randomness but remains above the deterministic baseline.

Given remaining gaps with human strategies, we conduct qualitative studies (Fig.~\ref{fig:vis drivelm}) in representative scenarios. (a) With all surrounding vehicles stationary, the system drives fast for efficiency, collapsing to a unimodal strategy matching ground-truth. (b) A red light and leading vehicle's behavior prompt slow driving; though not covering the ground-truth (fast driving), both options are interpretable and rule-compliant. (c) At an intersection with pedestrians and construction barriers, the system recommends remaining stationary (0.903) due to pedestrian proximity, matching ground-truth; a low-speed left turn is also comprehensible as pedestrians diverge from the ego path. (d) At night with low visibility and a right-turn navigation, the decision bifurcates between right-turn (0.802) and straight (0.198), with the latter matching ground-truth. This deviation represents a safety-prioritized alternative.

In summary, while generated multimodal strategies do not perfectly align with human behaviors, they remain semantically plausible even when discrepancies occur.

\begin{table}[t]
    \centering
    \begin{minipage}{\columnwidth}
        \caption{Open-loop performance of the multi-agent analytical system under different action voting temperatures ($\widetilde{\tau}$).}
        \label{tab:temperature scaling}
    \end{minipage}

    \vspace{-0.2cm}
    
    \resizebox{0.4\textwidth}{!}{%
    \begin{threeparttable}
    \begin{tabular}{@{}p{0.18\columnwidth}p{0.12\columnwidth}M{0.17\columnwidth}M{0.17\columnwidth}M{0.17\columnwidth}@{}}
        \toprule
        \textbf{Model} & \textbf{$\widetilde{\tau}$} & \textbf{Acc-1 (\%)}  & \textbf{Acc-2 (\%)}  & \textbf{Acc-3 (\%)}  \\
        \midrule
        \textcolor{mediumgray}{A.V.\,-\,Only} & \textcolor{mediumgray}{0.0} & \textcolor{mediumgray}{59.67} & \textcolor{mediumgray}{-} & \textcolor{mediumgray}{-}\\
        \midrule
        \multirow{5}{*}{M.A.A.S$^\dagger$} & 0.3 & 63.00 & 70.67 & 72.67\\
        & 0.5 & 64.67 & 72.67 & 75.33 \\
        & 0.7 & \makecell{\colorbox{lightblue}{65.00}} & 74.00 & \makecell{\colorbox{lightblue}{78.33}} \\
        & 0.9 & 63.33 & 72.67 & 76.33 \\
        & 1.1 & 63.67 & \makecell{\colorbox{lightblue}{74.67}} & 77.67 \\
        \bottomrule
    \end{tabular}
    \begin{tablenotes}
        \item[$\dagger$]: Multi-agent analytical system. 
    \end{tablenotes}
    \end{threeparttable}
    }
    \vspace{-0.5cm}
\end{table}

\begin{figure*}
  \centering
    \includegraphics[
    width=\linewidth]{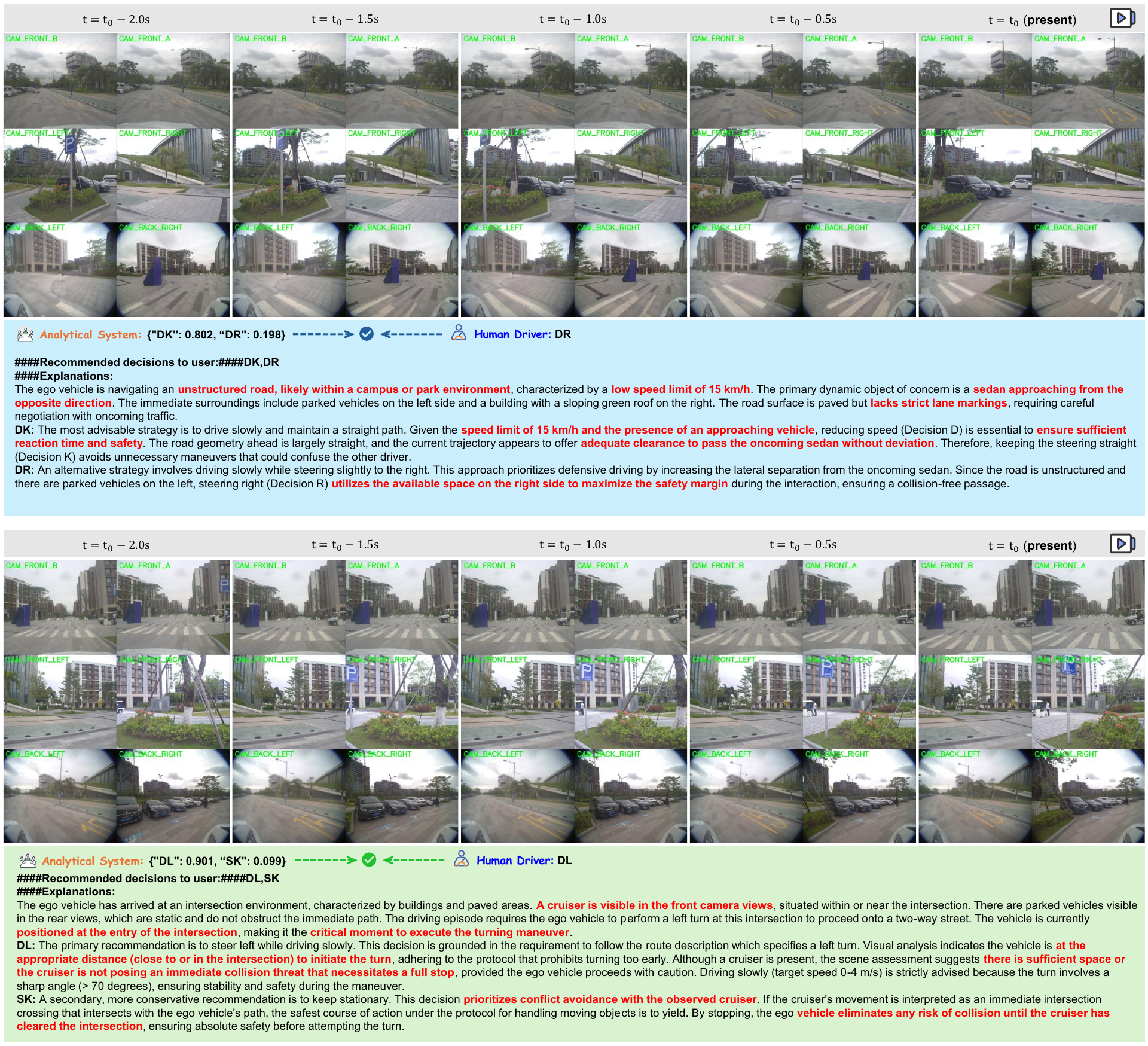}
  \caption{Representative cases from our in-house campus dataset for demonstrating the analytical system's ability to perform multimodal decision reasoning with multi-view video inputs. The definitions of symbols and action codes are consistent with Fig.~\ref{fig:vis drivelm}. Key contents in responses are highlighted in \textcolor{red}{red}.}
  \label{fig:vis hkustgz}
  \vspace{-0.3cm}
\end{figure*}

\textbf{Case Studies with Field-Collected Data.} Fig.~\ref{fig:vis hkustgz} showcases our analytical system's decision-making capability in a college campus environment with multi-frame history inputs.

In the first episode, the ego vehicle encounters a sedan emerging from a parking area on a wide, unstructured road. Without visible lane delineations, the system prioritizes a straight slow path (DK, 0.802) due to adequate clearance, while assigning a significant probability to a defensive right steer (DR, 0.198). Though the human driver chose DR, the system's secondary mode covers this action, showing flexibility in allocating safety margins within unstructured spaces.

In the second episode, the ego vehicle approaches a left-turn intersection with a cruiser nearby. The system prioritizes a slow left turn (DL, 0.901) and assigns a minor probability to stopping (SK, 0.099), validating confident execution of commanded maneuvers with a safety-oriented fallback.

Collectively, these episodes demonstrate the system's capacity to adaptively modulate uncertainty, offering diverse safe alternatives in ambiguous scenarios while maintaining decisiveness in structured contexts.

\section{Conclusion}
In this paper, we introduce LUNA-AD, a comprehensive framework designed to reconcile complex multimodal decision reasoning with the efficiency requirements of onboard AD systems. Through our proposed tri-system architecture, we successfully disentangle deliberative analytical reasoning from lightweight heuristic inference, enabling the preservation of uncertainty semantics without compromising inference efficiency. The integration of uncertainty-aware distillation and a reflection-driven lifelong learning mechanism allows the system to continuously adapt to distribution shifts and closed-loop feedback. Empirical evaluations on the nuPlan benchmarks substantiate that LUNA-AD achieves SOTA success rates while maintaining strong robustness, significantly reducing computational latency compared to existing large foundation model-based approaches.

Future work will aim to evolve the current structured state-based framework toward direct vision-language integration. While our preliminary experiments on DriveLM-NuScenes and field-collected data have validated the feasibility of the analytical system with raw RGB inputs, future improvements will focus on end-to-end perception-reasoning pipelines that eliminate reliance on intermediate state descriptors, thereby enhancing generalization in unstructured environments. 
\bibliographystyle{ieeetr}
\bibliography{main}

\end{document}